\documentclass{article}

\usepackage[preprint]{corl_2026}
\usepackage{xspace}
\usepackage{amsmath, amssymb}
\usepackage{amsthm}
\usepackage{mathtools}
\usepackage{algorithm}
\usepackage{algpseudocode}
\usepackage{siunitx}
\algnewcommand{\RequireCont}{\item[\hphantom{\algorithmicrequire}]}%
\makeatother
\usepackage{cleveref}
\usepackage{svg}
\usepackage{bbding}
\usepackage{booktabs}
\usepackage{fancyvrb}
\usepackage{multirow}
\usepackage{bm}
\usepackage[section]{placeins}
\usepackage{subcaption}
\usepackage[export]{adjustbox}

\usepackage{etoc}
\usepackage{xstring}

\definecolor{commentgray}{RGB}{100,100,200}
\algrenewcommand\algorithmiccomment[1]{\hfill{\color{commentgray}\(\triangleright\) #1}}
\newcommand{\mypara}[1]{\par\noindent\textbf{#1}\ }

\usepackage[T1]{fontenc}
\usepackage{times}

\DeclareMathOperator*{\argmax}{arg\,max}

\newcommand{\name}{MPAIL2\xspace}
\newcommand{\seq}[1]{\mathbf{#1}}

\newcommand{\zspc}{\mathcal{Z}}
\newcommand{\aspc}{\mathcal{A}}

\newcommand{\expd}{\mathcal{D}}

\newcommand{\namep}{[$-$P]\xspace}
\newcommand{\nameo}{[$-$O]\xspace}
\newcommand{\namepm}{[$-$PM]\xspace}
\newcommand{\namepmo}{[$-$PMO]\xspace}

\newcommand{\R}{\mathbb{R}}
\newcommand{\enc}{e_\omega}
\newcommand{\dyn}{f_\psi}
\newcommand{\rew}{r_\theta}
\newcommand{\val}{Q_\zeta}
\newcommand{\pol}{\pi_\phi}
\newcommand{\buf}{\mathcal{B}}

\newcommand{\targ}

\newcommand{\addmpail}[1]{\textcolor{blue}{#1}}

\newcommand{\claim}[1]{\textbf{#1}}
\newcommand{\website}{\href{https://uwrobotlearning.github.io/mpail2/}{https://uwrobotlearning.github.io/mpail2/}}

\hypersetup{pdftitle={Online World Modeling Enables Real-World Inverse Reinforcement Learning from Observation},
  pdfauthor={Tyler Han, Bat Nemekhbold, Siyang Shen, Rohan Baijal, Richard Ebock, Harine Ravichandiran, Sanghun Jung, Kevin Huang, Byron Boots}}

\title{Online World Modeling Enables Real-World\\ Inverse Reinforcement Learning from Observation}

\author{
  Tyler Han\\
  \And
  Bat Nemekhbold\\
  \And
  Siyang Shen\\
  \And
  Rohan Baijal\\
  \And
  Richard Ebock\\
  \And
  Harine Ravichandiran\\
  \And
  Sanghun Jung\\
  \And
  Kevin Huang\\
  \And
  Byron Boots\\
  \AND
  \textmd{University of Washington}
}

\begin{document}\maketitle

\vspace{-24pt}
\begin{abstract}
Current methods in robot learning are fundamentally bottlenecked by one or more of: hand-designed rewards, simulation modeling, or action supervision (e.g. teleoperation) each requiring significant domain expertise, engineering effort, and robot-operator labor. Towards reducing these bottlenecks, this work pursues observational learning via Inverse Reinforcement Learning from Observation (IRLfO) in which only access to task observations (e.g. video) is assumed. Due to the challenging setting and limitations of RL methods, IRLfO has thus far remained impractical for real-world robot learning. Here, we present the first IRL method to learn visual manipulation in the real world from scratch, and the first real-world demonstration of positive online transfer across visual manipulation tasks from scratch. In under 40 minutes, MPAIL2 learns pick-and-place from scratch to 82\%, where RL and BC with equal interaction and demonstration budgets reach only 0\% and 12\% despite their reward and action supervision. Interactive project page with training videos: \website
\end{abstract}

\begin{figure*}[htb]
  \centering
  \includegraphics[width=0.95\linewidth,trim=0 8cm 0 0,clip]
  {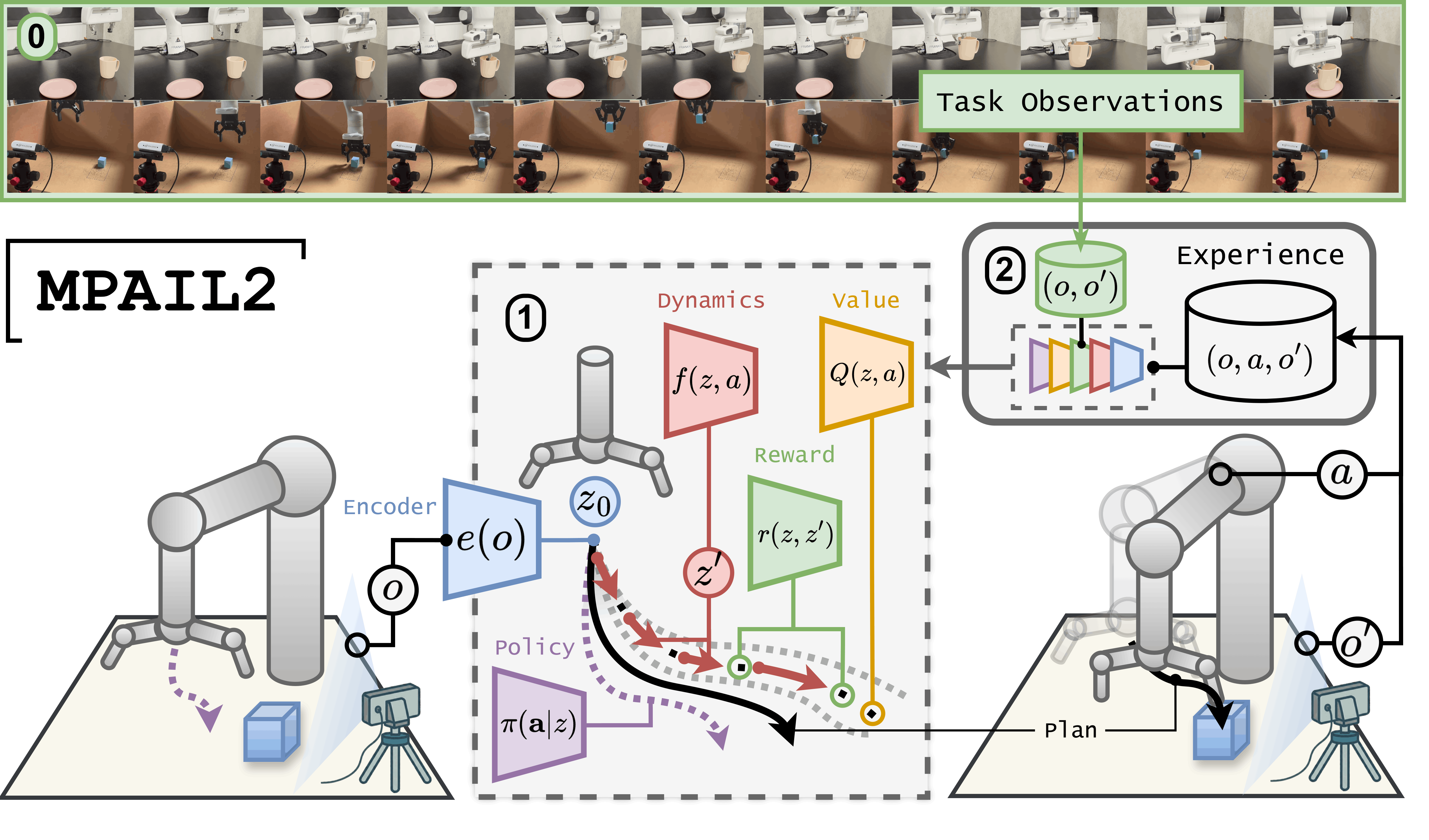}
  \caption{Overview of MPAIL2. \textbf{(0)} \textit{Only} task observations are provided to the learner. \textbf{(1)} The learner makes and \textcolor{CornflowerBlue}{\textbf{encodes}} observations into a latent state representation. The \textcolor{Orchid}{\textbf{policy}} predicts reactive actions which seed online planning. Performing online planning, the learner samples actions, predicts their \textcolor{BrickRed}{\textbf{future states}}, and evaluates plans using the online \textcolor{OliveGreen}{\textbf{reward}} and the \textcolor{orange}{\textbf{value}}. Plans with higher returns are more likely to contribute to the executed action. \textbf{(2)} After executing, the interaction $(o,a,o')$ where $o'$ is the posterior observation is accumulated in the learner's experience. All components are updated online.}
  \label{fig:fig1}
\end{figure*}

\section{Introduction}\label{sec:intro}

You are learning to receive a baseball pitch for the first time. Before you, your coach demonstrates the task by receiving then striking the ball. You step to the plate and mimic what you observed, but as the pitch arrives, you begin to experience the challenges your coach has long mastered: the sudden inertia of the bat, the shifting balance in your feet, and the ball's inbound trajectory. Even when provided a demonstration, we are often not immediately capable of completing a task without firsthand experience. However, our ability to infer intent from a few observations, self-evaluate, and improve in real-time sets us apart from robots today.

Experiential improvement on robots today is far from self-supervised: pre-training, prior models, simulation, manual data curation, reward labeling, failure examples, and tele-operated interventions are some examples of manual bottlenecks used to combat the poor sample efficiency of real-world Reinforcement Learning (RL). Reliance on extensive manual supervision weakens the applicability of RL and distances the approach from the deployment setting, where expert operators may be absent and interaction is most diverse and valuable for improvement.

Meanwhile, demonstration data may be currently underutilized in RL pipelines, which often employ direct supervision, limiting coverage and generalization. Alternatively, interactive learning of a reward to perform \textit{indirect} supervision has largely been impractical for real-world robot learning. Known broadly as Inverse Reinforcement Learning (IRL), we hypothesize that its theorized gains in generalization have been confounded with losses in instability caused by an adversarial reward.

By approaching IRL through online world modeling, we combat this instability through three useful properties:
Latent \textbf{Predictability} forces adversarial changes in reward to remain predictable throughout training.
\textbf{Propagation} via multi-step return estimation enables direct optimization of a globally changing reward.
Actor \textbf{Robustness} via online planning mitigates policy sub-optimality through broad action re-optimization.

We find that all three properties are required for practical, real-world IRL from Observation (IRLfO).
To our knowledge, this work is the first demonstration of visual manipulation like pushing and pick-and-place from scratch\footnote{\textit{from scratch}: no pre-training used anywhere; this includes encoders, policies, critics, etc.} in under 40 minutes. SOTA RL and Behavior Cloning baselines reach only 0\% and 12\% success, respectively, at matched interaction budgets, even with rewards and action labels.

Our main contributions are as follows:

\begin{enumerate}
\item \name, an online, latent world-modeling framework, which simultaneously improves encoder, dynamics, reward, value, and policy models using only observations and interaction. Towards learning in deployment settings, \textit{no rewards, simulation, pre-training, or action supervision are assumed.}
\item Over 96 independent real-world, training-from-scratch experiments validating and isolating designs through ablations in the real world. Due to significant reduction in real-world training-from-scratch times and reduction of online supervision (e.g. interventions, rewards, labeling), the following experimental analyses also benefit from one of the most comprehensive multi-seed evaluations in real-world RL.
\item Experiments toward scaling robot learning continually in the real-world beyond single-task imitation, through positive online transfer learning and video-only demonstrations via \name. To our knowledge, this work is also the first to demonstrate positive transfer learning entirely in the real world without offline pre-training.
\end{enumerate}

We provide a closing discussion with auxiliary experiments to investigate why the proposed properties interact favorably and how they might be generalized for future work.
Training videos, code, and more can be found on the interactive project website: \website.

\section{Related Work}

\textbf{Learning from Observation (LfO).} Methods in LfO learn to perform tasks from observation alone, theoretically free of hand-designed rewards and teleoperation. In practice, however, effective LfO approaches can require vast amounts of prior data, direct action supervision (e.g. teleoperation), or hours of demonstrations for seconds-long tasks~\cite{liu_imitation_2018, wang_mimicplay_2023, cui_play_2022, rosete-beas_latent_2023}. These methods often produce large, unwieldy models relying on domain knowledge and embodiment-specific data~\cite{wang_mimicplay_2023, open_x-embodiment_collaboration_open_2024}, with domain-specific representations that do not readily generalize to experiential improvement via Reinforcement Learning.

\mypara{Inverse Reinforcement Learning from Observation (IRLfO)} is known to drastically~\cite{han_model_2025, jain_smooth_2025} reduce demonstration quantity by trading off direct-supervision via demonstrations for self-supervision via RL-like exploration and interaction~\cite{abbeel_apprenticeship_2004, ho_generative_2016, finn_guided_2016}. However, needing to simultaneously optimize for both a reward function and a policy, IRL methods depend upon and exacerbate the already poor real-world interaction efficiency of existing RL algorithms. As a result, IRL approaches currently can require impractical amounts of online interaction on the order of thousands of policy rollouts~\cite{sun_adversarial_2021, han_model_2025, das_model-based_2021, finn_guided_2016, torabi_sample-efficient_2019}.
To the best of our knowledge, few works have investigated real-world IRL for robotics and none that forgo action supervision (IRLfO). Notable is the model-based IRL work in~\cite{sun_adversarial_2021} demonstrating visual reaching in the real-world. Other works in the domain rely on prior modeling and pre-trained visual features such as ~\cite{finn_guided_2016, das_model-based_2021, abbeel_apprenticeship_2004, ratliff_learning_2009}.

\mypara{Real-world Reinforcement Learning.}
Popular methods in real-world RL instantiate learning as RL or RL with Prior Data (RLPD)~\cite{ball_efficient_2023, luo_serl_2025, luo_precise_2025, hansen_modem_2022,  lancaster_modem-v2_2024, lei_performant_2026}.
In RL or RLPD, supervision signal arrives in the form of scalar rewards. The learner is also supervised directly at the state-action level, through offline BC pre-training, interventions, and symmetric sampling. In these frameworks, rewards can also be learned interactively as in~\cite{fu_variational_2018}. However, all prior work, to our knowledge, does not bring the reward online. Indeed, we find that model-free approaches under the adversarial reward suffer greatly.

\section{Model Predictive Adversarial Imitation Learning 2}\label{sec:method}

\mypara{Approach.}
The \textit{task-agnostic} latent representation is self-supervised through the encoder $z_t = \enc(o_t)$ and latent dynamics $\hat{z}_{t+1} = \dyn(z_t, a_t)$~\cite{grill_bootstrap_2020}. The inferred reward $r_t = \rew(z_t, z_{t+1})$ and value $q_t = \val(z_t, a_t)$ are used for offline, off-policy model-based policy optimization as well as for evaluating sampled plans during online planning. The multi-step policy $\hat{\seq{a}}_{t:t+H} \sim \pol(\cdot \mid z_t)$ produces $H$-step plans that warm-start planning and support off-policy value estimation. Learner interactions $(o_t, a_t, o_{t+1})$ are accumulated in a replay buffer $\mathcal{B}$. Figure~\ref{fig:fig1} visualizes one interaction loop. As outlined in~\Cref{alg:mpail2} (\Cref{app:algorithms}), each of the above components are optimized independently in each learner update. The remainder of \Cref{sec:method} provides details on these components and their optimization objectives.

\mypara{Problem.}
We consider a Markov Decision Process (MDP) $(\mathcal{O}, \mathcal{A}, T, R, \gamma)$ whose states are the robot's observations $o \in \mathcal{O}$. The transition reward $R$, dynamics $T$, and terminal states are unknown.
The learner receives a set of observation trajectories $\mathcal{D} = \{(o_1, \dots, o_{T_i})_i\}_{i=1}^{N}$ generated by an optimal expert policy $\pi_E$. No action or reward supervision is provided. 
Following the apprenticeship-learning view of IRL \cite{abbeel_apprenticeship_2004}, the learner seeks to infer candidate rewards $r\in\mathcal{R}$ and policies $\pi\in\Pi$ by assuming the expert performs optimally in expectation:
\begin{equation}
\arg\max_{r \in \mathcal{R}}\;
\mathbb{E}_{\pi_E}\!\left[\,r\,\right]
-
\max_{\pi \in \Pi}\;
\mathbb{E}_{\pi}\!\left[\,r\,\right]
\label{eq:apprenticeship}
\end{equation}
While MPAIL demonstrates that online planning significantly improves learner robustness in IRL, it is algorithmically limited by its state-based dynamics, on-policy reward and value optimization~\cite{ho_generative_2016, kostrikov_discriminator-actor-critic_2018}, leading to reliance on prior modeling as well as massively simulated interactions~\cite{han_model_2025}.

\mypara{Notation.} Hats denote model predictions, bars slow target networks, and bold an $H$-step plan $\seq{a}_t = a_{t:t+H}$. A plan is rolled out from $\hat z_t \coloneqq z_t = \enc(o_t)$ by $\hat z_{t'+1} = \dyn(\hat z_{t'}, a_{t'})$, giving predicted rewards $\hat r_{t'}$ and values $\hat q_{t'}$. Logged transitions $(z, a, z') \sim \enc(\buf)$ and demonstration pairs $(z, z') \sim \enc(\expd)$ are encoded samples.

\mypara{Encoder \& Dynamics}\label{sec:method-dyn}
The goal of the latent dynamics are two-fold: (1) to enable online planning without requiring predictions directly in high-dimensional observation space and (2) to anchor task (reward) learning by restricting discrimination to lie within representations that are supported by the learner's direct interactions, i.e. its world model. 
Thus, the encoder and dynamics are trained jointly, task-agnostic, and self-supervised, by minimizing the
discounted multi-step latent prediction loss
\begin{equation}
\mathcal{L}_{e, f}(\omega, \psi)
= \mathbb{E}_{(o_{t:t+H},\,a_{t:t+H}) \sim \buf}\!\left[
\sum_{t' = t}^{t + H} \rho^{\,t' - t}\,
\big\|\hat{z}_{t'} - \mathrm{sg}(z_{t'})\big\|_2^2
\right] ,
\label{eq:enc-dyn-loss}
\end{equation}
where $\hat z_{t'}$ is rolled out along the logged actions, $\rho \in (0, 1)$ is a temporal discount and $\mathrm{sg}(\cdot)$
stops gradients on the target latent to prevent representational
collapse~\cite{chen_exploring_2021, tang_understanding_2023}. The loss
depends only on observed transitions, independent of any reward or
value head, yielding a task-independent representation that supports
transfer (\cref{sec:q3}).

\mypara{Inferred Reward}\label{sec:method-reward} Besides improved sample efficiency, training the reward off-policy maintains its coverage over all previous interactions, helping to stabilize the learned reward signal while generalizing it to more states~\cite{kostrikov_discriminator-actor-critic_2018, zhang_rewind_2025}. Otherwise, on-policy reward learning converges meaninglessly to a constant~\cite{fu_learning_2018}. Greater reward coverage is also necessary for reliable sample-based planning, in which the reward must evaluate a number of sampled states online that may not necessarily be in-distribution. The reward's training objective realizes the top-level apprenticeship-learning problem in \Cref{eq:apprenticeship} by latent world model surrogacy,
\begin{equation}
\arg\max_{r \in \mathcal{R}}\;
\mathbb{E}_{(z, z') \sim e_\omega(\mathcal{D})}\!\left[r(z, z')\right]
- \mathbb{E}_{(z, z') \sim e_\omega(\mathcal{B})}\!\left[r(z, z')\right]
- \beta \cdot \mathrm{GP}(r, \mathcal{B}, \mathcal{D}) \,,
\label{eq:reward-obj}
\end{equation}
where non-expert policies of Equation~\eqref{eq:apprenticeship} are
represented by samples from $\mathcal{B}$ and reward candidacy $r\in\mathcal{R}$ is regularized by the gradient
penalty~\cite{gulrajani_improved_2017} (defined in \Cref{app:wasserstein-gp}).
\mypara{Value}\label{sec:method-value}
The value supports recovery behavior learning, demonstration generalization, enables reasoning beyond the planning horizon, and provides a stable critic for the policy update. A $Q$-ensemble is used for variance reduction. The training objective is the
entropy-regularized off-policy $\lambda$-return target~\cite{haarnoja_soft_2018}
\begin{equation}
\mathcal{L}_Q(\zeta)
= \mathbb{E}_{(z, a, z') \sim \enc(\buf),\; \hat{\mathbf{a}}' \sim \pi_\phi(\cdot \mid z')}\!\left[
\left(Q_\zeta(z, a) - \left[\rew(z,z') + \gamma\,\bar{G}^\lambda(z', \hat{\mathbf{a}}')\right]\right)^2
\right] ,
\label{eq:value-loss}
\end{equation}
with $\bar{G}^\lambda(z', \hat{\mathbf{a}}')$ the $\lambda$-return of the policy's plan from $z'$, computed from the slow
target $\bar{Q}_\zeta$ via \Cref{eq:lam-return}. Polyak updates~\cite{haarnoja_soft_2018}
and ensembling~\cite{chen_randomized_2020} extend standard
latent-planning practice from the known-reward
setting~\cite{hansen_td-mpc2_2023} to our inferred-reward case.

\mypara{Multi-step Policy}\label{sec:method-policy}
We train $\pi_\phi$ to seed the planner with $H$-step warm-start plans. 
Single-step policies in current planning-based learners~\cite{hansen_td-mpc2_2023} can otherwise induce computationally expensive rollouts during online planning, requiring several inferences from policy-then-dynamics in sequence.
Also unlike multi-step policies supervised directly by expert action labels~\cite{zhao_learning_2023, chi_diffusion_2025}, the IRLfO setting has no action labels available. Supervision arrives entirely through the inferred reward and value. The policy network is optimized against a TD($\lambda$)-style return over $H$-step imagined rollouts with entropy regularization,
\begin{equation}
\max_{\pi \in \Pi}\;
\mathbb{E}_{\hat{\mathbf{a}}_t \sim \pi(\cdot \mid z_t)}\!\left[
G^\lambda(z_t, \hat{\mathbf{a}}_t)
- \alpha\, \log \pi(\hat{\mathbf{a}}_t \mid z_t)
\right] ,
\label{eq:policy-obj}
\end{equation}
where the $\lambda$-return mixes the bootstrap $\hat q_t$ with a TD($\lambda$) return $\hat G_t$ rolled out through the learned dynamics,
\begin{equation}
G^\lambda(z_t, \seq{a}_t) \coloneqq \lambda\,\hat q_t + (1-\lambda)\,\hat G_t\,,\qquad
\hat G_{t'} \coloneqq \hat r_{t'} + \gamma\left[(1-\lambda)\,\hat q_{t'+1} + \lambda\,\hat G_{t'+1}\right],
\label{eq:lam-return}
\end{equation}
for $t \le t' < t+H-1$, with $\hat G_{t+H-1} \coloneqq \hat q_{t+H-1}$; $\hat G_t$ is the standard TD($\lambda$) return~\cite{sutton_reinforcement_2018}.
The temperature $\alpha$ is the SAC-style entropy coefficient~\cite{haarnoja_soft_2018}, tuned to a target entropy per action step. As we find in this work, the policy plays a comparatively small role in decision making relative to the planner, acting primarily to seed the planner and support off-policy value estimation (\Cref{fig:pol-influence}).

\mypara{Planner}\label{sec:method-planner}
As previously found, planning-based actors exhibit significantly more robust and generalizable behavior compared to policy-based actors~\cite{han_model_2025, hansen_td-mpc2_2023, jain_smooth_2025}.
Here, we choose Model Predictive Path Integral (MPPI)~\cite{williams_information-theoretic_2018} due to its proven theoretical connections~\cite{han_model_2025}, simplicity, extensibility~\cite{pan_sampling-based_2025}, popularity, and familiarity in other planning-based learners~\cite{han_model_2025, jain_smooth_2025, hansen_td-mpc2_2023}.
A small fraction of the
sampled plans is drawn from $\pi_\phi(\cdot \mid z_t)$ to warm-start
MPPI, and the remainder is sampled from the current proposal.

\mypara{Algorithm and Architectures.} Algorithmic and architectural details of \name's models, training, and planning procedures can be found in \Cref{app:algorithms} and \Cref{app:architecture}, respectively. We encourage the reader to view our interactive project page, which systematically motivates and builds up to each of these components: \website. \Cref{app:related-designs} contrasts \name's task-agnostic latent with the task-shaped latent of TD-MPC2~\cite{hansen_td-mpc2_2023}, and its multi-step policy with action chunking~\cite{zhao_learning_2023}.

\section{Experiments}\label{sec:exp}
Our experiments are motivated by the following research questions:
\begin{enumerate}
    \item[\textbf{\ref{sec:q1}}] \textbf{Enabling Real-world IRLfO.} How does the design of \name enable real-world IRLfO?
    \item[\textbf{\ref{sec:q2}}] \textbf{IRLfO vs. RL/BC.} Does less direct supervision mean less performance?
    \item[\textbf{\ref{sec:q3}}] \textbf{Scaling real-world learning.} How does \name scale real-world robot learning beyond single-task imitation?
\end{enumerate}

\paragraph{Experiment Setup.}

All experiments use human-operated demonstrations~\cite{orsini_what_2021}. Reward and success definitions and the reset protocol are in \Cref{app:exp-details}. Results report two checkpoints: \textit{Best} (best of saved checkpoints) and \textit{Last} (final checkpoint), the latter quantifying performance when stopping criteria are unavailable. Best is selected post hoc among checkpoints saved every 10 episodes, identically for \name and every online baseline; success labels are used only for this selection and never during training. Reported training times include interaction, resets, and learner updates, and exclude checkpoint evaluations. %

Real-world tasks are evaluated on two independent robot platforms (Franka: Block Push and Mug-on-Plate; Kinova Gen3: Pick-and-Place). Tasks are described in~\Cref{tab:tasks} (\Cref{app:exp-details}).
\textit{In transfer learning}, methods are first trained on an initial task, e.g. Block Push, then on a new but related task, like pushing in the opposite ($+y$) direction, with model weights initialized from the first task.
Only \namep and BC are evaluated as they are the only other methods to achieve consistent success on the initial task.
\textit{In video-only demonstration tasks}, only the external, table-mounted camera is provided as observations. Action-supervised methods are still provided actions. Architectural changes made to \name for video-only experiments are detailed in~\Cref{app:exp-details-substate}.

\paragraph{Baselines.}

We compare with the following baselines, organized by their problem scope (\Cref{tab:results}).

\noindent\textit{IRL Methods.}
\textbf{AIRL}~\cite{fu_learning_2018}, \textbf{DAC}~\cite{kostrikov_discriminator-actor-critic_2018}, \textbf{MAIRL}~\cite{sun_adversarial_2021}, and \textbf{MPAIL}~\cite{han_model_2025} operate in the same IRLfO scope as \name and are trivially extended to the observation-only setting by learning rewards over latent state-transitions~\cite{torabi_generative_2019}.
Towards answering \textbf{Q1} and fair comparison, we improve upon each of these baselines by applying \name's design choices where possible. These designs are not part of the original proposals but have been empirically validated in real-world robot learning~\cite{ball_efficient_2023, luo_serl_2025}.
As a result, these IRL baselines can also be analyzed as ablations on \name's integration of planning [P], dynamics modeling [M], and off-policy optimization of reward and value [O]. Each baseline's components are summarized alongside their evaluation results in~\Cref{tab:results}.

\noindent\textit{Reinforcement Learning with Prior Data (RLPD).} \textbf{RLPD} is a state-of-the-art method in real-world RL-with-demonstrations. It operates at the scope where hand-designed reward and action supervision are available~\cite{ball_efficient_2023}. We compare \name (without reward or action access) to RLPD implemented via~\cite{luo_serl_2025} with dense reward where known and action-labeled demonstrations.

\noindent\textit{Behavior Cloning (Diffusion).}
\textbf{Diffusion Policy}~\cite{chi_diffusion_2025} demonstrates ideal, state-of-the-art performance when the environmental reset is ``in-distribution'' to the provided set of demonstrations. 

\renewcommand{\thesubsection}{Q\arabic{subsection}}
\subsection{World modeling \textit{and} planning together enable real-world IRLfO.}\label{sec:q1}

\begin{figure*}[t]
    \centering
    \includegraphics[width=\linewidth]{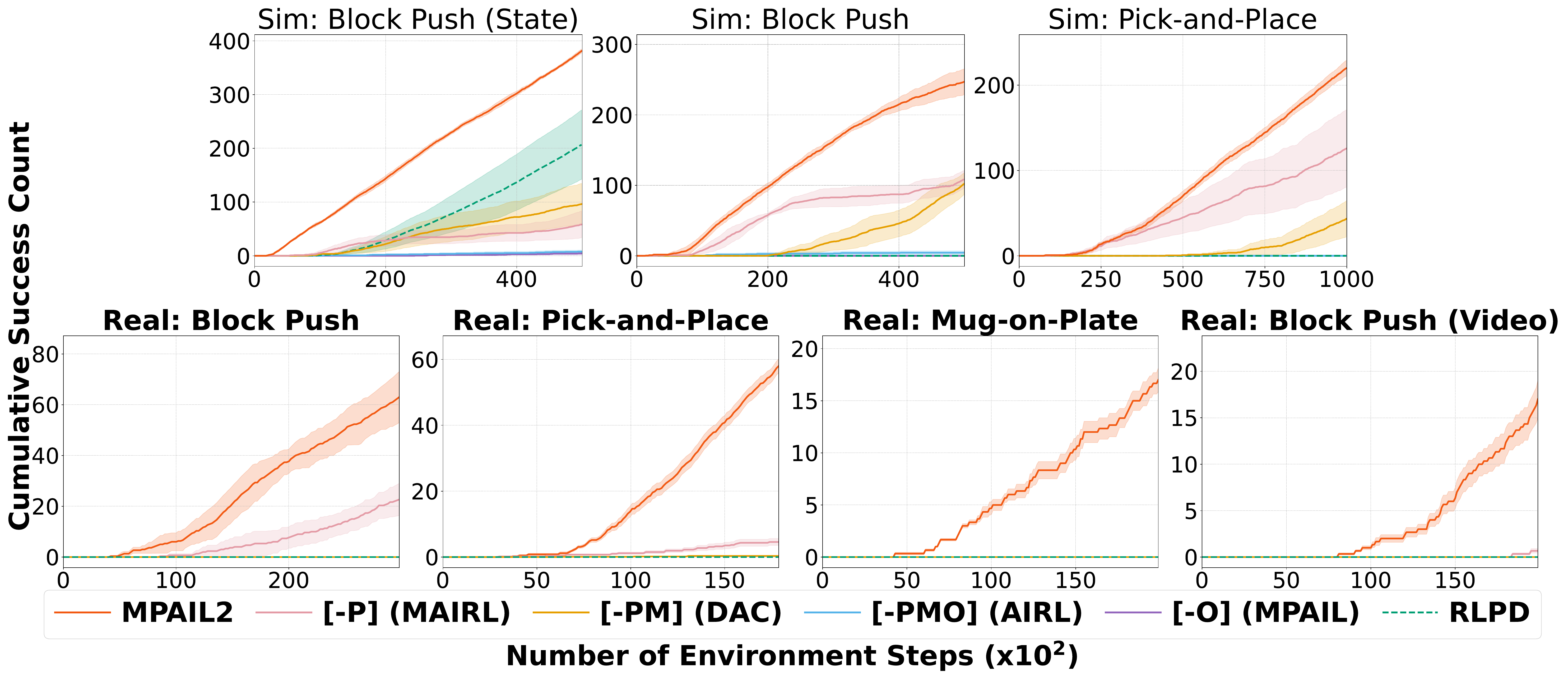}
    \caption{Cumulative successes over training across simulated and real experiments. The onset and slope of each curve correspond to the learning time and approximate success rate of the method. Evaluations by checkpoint reported in~\Cref{tab:sim-eval}. \namepmo (AIRL) and \nameo (MPAIL) are not evaluated in real due to their negligible success rates across simulation tasks.}
    \label{fig:sim-results}
\end{figure*}

We find that all of \name's design choices contribute positively towards alleviating
the sample efficiency and robustness challenges that have made IRLfO and its approaches historically impractical. We make two significant observations: 

\claim{World modeling is critical to efficiency in IRLfO.}
Of all evaluated IRL methods, those utilizing a latent dynamics model (\name and \namep) are the only approaches which demonstrate any successes when training in the real-world. Even in simulation, \name and \namep demonstrate successes significantly (2--3$\times$) earlier in training than other methods. We further observe that off-policy training is itself critical. \nameo, which preserves \name's world model, planner, and adversarial objective but trains the reward, value, and policy on-policy, fails to match \name's sample efficiency or stability under a matched interaction budget (\Cref{fig:sim-results} and \Cref{tab:sim-eval}).

\claim{Planning critically mitigates IRL instability and improves actor robustness, enabling real-world IRLfO.}
Simulation results reveal and corroborate expected adversarial instability~\cite{orsini_what_2021}, as reflected by the decaying or inconsistent slopes in \Cref{fig:sim-results} corresponding to approximate drops in success rate. However, we observe that the instability is substantially mitigated via planning. While \namep's success rate tends to drop without resurgence, \name exhibits little to no signs of instability with the exception of the late decay in Sim: Block Push. The impact of the planner's stability is clear in the real-world experimentation, where in many cases, \name is the only IRL method to demonstrate consistent improvement over the course of training. At its best checkpoints, it is the \textit{only} IRLfO method to demonstrate improvement over BC, falling below only at last checkpoints on Block Push where Diffusion is saturated. \name is actively expanding the robot's capability beyond the demonstration coverage strictly through interaction \textit{with neither rewards nor action supervision}.

\subsection{Comparing IRLfO, RL, and BC: less direct supervision does not mean less performance.}\label{sec:q2}

\textbf{\name and its IRLfO ablations significantly outperform or compete with the performance of RLPD and BC, although they receive strictly less supervision.}
We reiterate that: a) RLPD is provided action supervision and dense reward (or online hand-labeling for classifier-based reward~\cite{luo_serl_2025}); and b) BC is provided action supervision; none of this information is provided to the IRL methods. Interestingly, the minimal off-policy IRL ablation, \namepm (DAC), performs qualitatively better than RLPD by learning to reach within the allotted interactions (see project website for videos).
\namepm (DAC) differs from RLPD merely through reward type and action supervision. As a result, we see a distinct benefit of IRL through \textit{online, data-driven reward modeling}~\cite{arjovsky_wasserstein_2017}. These results support the notion that a learned reward, even one without negatively-labeled examples, can be more informative than a hand-designed one. 

Despite RLPD's overall poor performance in almost all evaluations, its performance in Block Push (state) is second only to \name and competitive in Gymnasium tasks (\Cref{fig:gym-envs}), validating its implementation. Under matched interaction budget, RLPD appears to scale poorly with observation dimensionality and degrades as manual supervision decreases, as opposed to \name, \namep (MAIRL), and even \namepm (DAC). Across all real-world runs, RLPD exhibits no qualitative improvement within the allotted interaction budget. Especially given RLPD's reliance on manual supervision, continuing to train RLPD to convergence is impractical for our multi-seed study. Additional discussion in~\Cref{app:discussion}.

\textbf{\name shows consistent improvement compared to BC by learning to generalize beyond the demonstrations through online interaction alone.}
Unlike supervised learning, which is unable to learn from failure or during deployment, new behavior can emerge autonomously under the (I)RL paradigm. For example, \name, like BC, tends to execute demonstrated motions even after missing a grasp in the first few minutes of training. After a few episodes (about 10 minutes), \name learns to re-attempt grasps when missed, before continuing the remainder of the demonstration. Unlike RL or RLPD, these behaviors are not directly supervised via rewards or negative demonstrations. Poor behaviors are implicitly de-incentivized in IRL merely through the principle that they partially resemble, or do not resemble at all, the observed demonstrations; and recovery behaviors necessarily emerge self-supervised through learned reward and value generalization.

\providecommand{\lst}[1]{{\color{gray}\footnotesize #1}} %
\begin{table*}[t]
\centering
\setlength{\tabcolsep}{2pt}
\setlength{\aboverulesep}{0pt}\setlength{\belowrulesep}{0pt}
\resizebox{\textwidth}{!}{\begin{tabular}{l cc|ccc| ccc >{\columncolor{gray!10}}c >{\columncolor{gray!10}}c >{\columncolor{gray!10}}c >{\columncolor{gray!10}}c >{\columncolor{gray!20}}c}
\toprule
& \multicolumn{2}{c|}{\textbf{Supervision}}
& \multicolumn{3}{c|}{\textbf{Components}}
& \multicolumn{3}{c}{}
& \multicolumn{2}{c}{\textbf{Transfer}}
& \multicolumn{2}{c}{\textbf{Scratch}} & \textbf{Video} \\
\cmidrule(lr){2-3}
\cmidrule(lr){4-6}
\cmidrule(lr){10-11}
\cmidrule(lr){12-13}
\cmidrule(lr){14-14}
\textbf{Method}
& Reward & Actions
& Plan. & Model & \shortstack{Off-\\Policy}
& BP & PnP & MoP
& BP (rt.) & PnP (rt.)
& BP (rt.) & PnP (rt.) & BP \\
\midrule
\textbf{\name}
  & & & \CheckmarkBold & \CheckmarkBold & \CheckmarkBold
  & \textbf{100$\pm$0} & \textbf{82$\pm$12} & \textbf{55 $\pm$ 15}
  & \textbf{90$\pm$5} & \textbf{94$\pm$10}
  & \textbf{88$\pm$6} & \textbf{53$\pm$37} & \textbf{63$\pm$10} \\
\lst{\quad last} & & & & & 
  & \lst{62$\pm$14} & \lst{68$\pm$14} & \lst{35$\pm$17} & \lst{64$\pm$4} & \lst{47$\pm$29} & \lst{80$\pm$12} & \lst{43$\pm$29} & \lst{59$\pm$14} \\
{[$-$P] (MAIRL)}
  & & & & \CheckmarkBold & \CheckmarkBold
  & 64$\pm$18 & 16$\pm$15 & 3.9 $\pm$ 2.8
  & 14$\pm$13 & 16$\pm$18
  & 53$\pm$12 & 12$\pm$12 & 29$\pm$8 \\
\lst{\quad last} & & & & & 
  & \lst{34$\pm$18} & \lst{16$\pm$15} & \lst{2$\pm$3} & \lst{8$\pm$6} & \lst{0$\pm$0} & \lst{34$\pm$10} & \lst{6$\pm$6} & \lst{26$\pm$12} \\
{[$-$PM] (DAC)}
  & & & & & \CheckmarkBold
  & 0$\pm$0 & 0$\pm$0 & 0$\pm$0
  & 0$\pm$0 & 0$\pm$0
  & 0$\pm$0 & 0$\pm$0 & 0$\pm$0\\
\midrule
BC (Diffusion)
  & & \CheckmarkBold & & &
  & \textbf{94$\pm$6} & 12$\pm$6 &  34$\pm$12
  & 16$\pm$8 & 18$\pm$6
  & 26$\pm$10 & 12$\pm$6 & 26$\pm$6 \\
RLPD
  & \CheckmarkBold* & \CheckmarkBold & & & \CheckmarkBold
  & 0$\pm$0 & 0$\pm$0 & 0 $\pm$ 0
  & 0$\pm$0 & 0$\pm$0
  & 0$\pm$0 & 0$\pm$0 & 0$\pm$0 \\
\bottomrule
\end{tabular}}
\caption{\textbf{Real-world task evaluations summarizing over 96 total training runs from scratch.} (3 seeds/task; 50 trials/task):
Success rate (\%, mean $\pm$ std across seeds) at best checkpoint and, in the small rows, at the last checkpoint; [$-$PM] and RLPD are 0 at every checkpoint (full checkpoint evaluations in~\Cref{app:ckpt-eval}). %
\namepmo (AIRL) and \nameo (MPAIL) are not evaluated in real experiments due to impractically poor sample efficiency.
\textit{Transfer}: new push/place direction, initialized from last checkpoint of initial task.
\textit{Scratch}: new task trained from scratch.
BP: Block Push, PnP: Pick-and-Place, MoP: Mug-on-Plate, rt.: right (opposite direction). *RLPD given known dense reward where available, otherwise classifier-based~\cite{luo_serl_2025} reward.
}
\label{tab:results}
\vspace{-12pt}
\end{table*}

\subsection{Scaling beyond single-task imitation: transfer learning and video-only demonstration.}\label{sec:q3}

Real-world robot learning is bottlenecked not only by per-task sample
efficiency, but by how training, supervision effort, and learned structure
scale across tasks.
The ability to accumulate, understand, and transfer experience throughout robot deployment autonomously is a long-standing promise of model-based methods and RL~\cite{sutton_reinforcement_2018}. Towards scalable real-world robot learning, we show that: \name (1) exhibits consistent positive transfer to new but related tasks, and (2) can learn from video-only demonstration.

\claim{\name learns significantly faster with weights transferred than trained-from-scratch, indicating positive representation transfer.} When training Block Push (rt.) from scratch, \name achieves the highest performing model (88\% success rate) after 140 episodes (50 minutes). When using a transferred model, only 60 episodes (20 minutes) are needed for the highest performing model (90\%). For pick-and-place, this transfer reduction is about 20 minutes \textit{while achieving roughly double the success rate of the best from-scratch model (53\% vs. 94\%)}.
Object segmentation, identification, and rigid-body dynamics are the primary dynamics of our evaluation tasks and plausibly the primary features undergoing transfer, evidenced by how \name spends much of the beginning learning to reach the object. Transfer is also supported by the following observation:

\claim{Planning mitigates negative transfer of a policy-based actor.}
BC and \namep exhibit negative transfer, corroborating existing work in literature~\cite{lyle_understanding_2022}. Due to the identical weight transfer procedure as \namep, \name's policy must likewise suffer from this phenomenon. Meanwhile, the planning-based actor nonetheless demonstrates positive transfer, suggesting that it may be enabled by disregarding policy output in cases of misalignment with reward and value during interaction.

\claim{\name can learn from scalable demonstration modalities (e.g. video-only).}
An advantage of IRLfO via \name is that the reward can be conditioned on only sensing modalities accessibly shared by both demonstrator and learner~\cite{zakka_xirl_2022}. For example, conditioning the reward model on only the external camera view may enable leveraging more accessible demonstration data, rather than restrictive combinations of wrist cameras, proprioception, and end-effector actions required by supervised learning~\cite{open_x-embodiment_collaboration_open_2024}. As a first step toward scalable third-person demonstration modalities, we evaluate learning from the fixed external camera only. With \name, the robot learns to perform the Block Push task from scratch to 63\% success rate using only demonstration observations from the table camera view, which is more than double BC's success rate (26\%) \textit{even with action supervision}. Experiment details in \Cref{app:exp-details-substate}.

\begin{figure}[t]
    \centering
    \includegraphics[width=\linewidth]{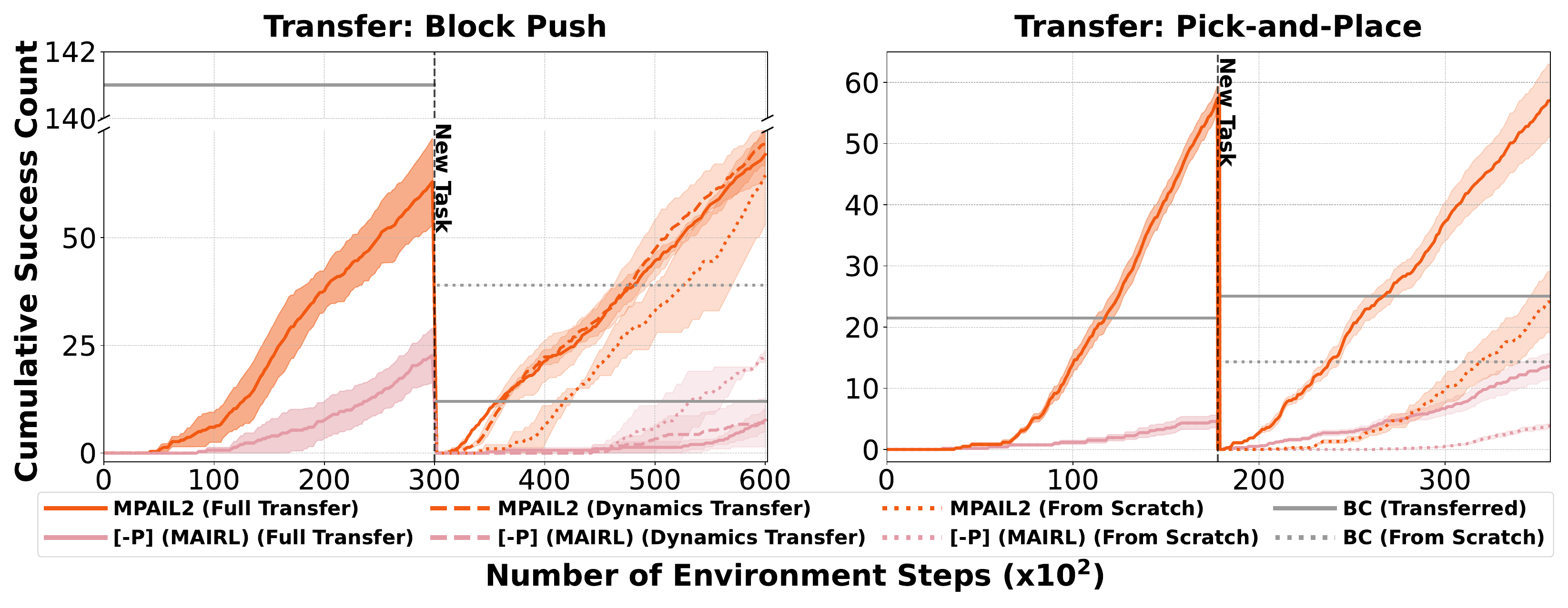}
    \caption{Cumulative successes in real-world transfer experiments compared to training from scratch. Methods are trained on the initial task (e.g. Block Push), then weights are transferred to training on a new task (e.g. Block Push right). \textit{Dynamics Transfer} reflects transfer of only encoder and dynamics model weights. Numerical results are in~\Cref{tab:real-eval-summary}.}
    \vspace{-12pt}
    \label{fig:transfer-results}
\end{figure}

\renewcommand{\thesubsection}{\thesection.\arabic{subsection}}

\section{Discussion}\label{sec:discussion}

We study \name's three properties with comparative analyses (\Cref{app:sim-ablations,app:world-model-figs}) to provide further insight into key questions and limitations for future work.

\mypara{Predictability.} When the reward and value losses also train \name's encoder, as in TD-MPC2, the latent collapses in four of five seeds; while collapsed, the reward cannot tell demonstrations from the robot's own data, and cumulative successes fall by more than half. On held-out real Block Push episodes, the task-agnostic world model predicts its own reward and responds to the action, though only modestly (\Cref{fig:wm}). Which task structure a task-agnostic latent should carry remains open.

\mypara{Propagation.} The learned reward changes quickly early in training and then settles. While it changes, the policy falls behind the best plan under the current reward, and further the faster the reward moves. How much multi-step return is needed to keep up remains open.

\mypara{Robustness.} Early in training, executing the planner succeeds far more often than executing the policy alone, by about 30 percentage points in every seed. Once the reward settles the order flips: the policy, trained on the planner's data, acts better than the planner, whose averaged output drifts from its own best sample. Planning thus matters most while the reward changes, suggesting planners that hand control to the policy, or sharpen their search, later in training.

\section{Conclusion} 
\label{sec:conclusion}

Due to its historically impractical learning efficiency and robustness, prior work in real-world IRL for robotics is scarce, especially so in the observation-only setting. In this work, we introduce \name to enable real-world IRLfO, demonstrated through the learning of visual manipulation tasks from scratch in under 40 minutes from observation \textit{without reward, action supervision, simulation, or pre-training}. In enabling real-world IRLfO, we find that the performance of \name and even its ablated IRL baselines does not suffer from less supervision and prior information compared to RL and BC. In fact, \name outperforms its RL and BC baselines by learning to generalize through self-guided online interaction. Our analyses tie this to its design: a task-agnostic latent avoids the collapse seen when task losses shape it, and online planning matters most while the learned reward is still changing. Beyond learning efficiency, enabling robots to learn and scale to tasks like baseball from observation will ultimately also depend upon supervision effort and transferability. To this end, we show that \name also enables positive transfer learning and learning from more scalable demonstration modalities. Through demonstrating these long-promised benefits of world modeling and IRL(fO) on over 96 real-world training runs from scratch, we hope this work encourages the study of the paradigm towards removing the bottlenecks of current robot learning.

\section{Limitations and Future Work}
\label{sec:future-work}

\mypara{Remaining supervision.} An operator still designates the task observations, resets objects between episodes, and selects the best checkpoint post hoc; safety comes from a constrained workspace, not the learner. Adversarial training can decay late (Sim: Block Push), so the final checkpoint can trail the best one (\Cref{tab:results}). Practical stopping criteria remain open.

\mypara{Scope.} All tasks are single-arm tabletop manipulation from $64{\times}64$ images; higher resolutions and longer horizons were not evaluated. The robot does not yet learn from other embodiments: the video-only demonstrations show the robot itself. Imagined rewards degrade with rollout depth (\Cref{app:world-model-figs}), bounding how far ahead plans can be trusted.

\mypara{Pre-training, prior data, world models, and cross-embodiment.}
Our transfer experiments reveal that all components of \name, even those that are task-dependent, may benefit from pre-training. Whether large pre-trained world~\cite{nvidia_cosmos_2025}, policy~\cite{black_pi_05_2025}, and reward~\cite{zhang_rewind_2025} models can be integrated with \name without giving up learning from the robot's own interaction remains open. Already, prior work in IRL has demonstrated cross-embodiment generalization when provided more data or via context translation~\cite{zakka_xirl_2022, zhang_peek_2025, hwang_masked_2026, liu_imitation_2018}.

\mypara{Recurrent state-based observation, probabilistic modeling, and recurrent dynamics.}
The present encoder architecture assumes that the latent state can be instantaneously observed. However, real-world robotics involves challenges such as hardware delay, PID tracking errors, simultaneous localization and mapping (SLAM) which involve information dynamics which persist through time. Recurrent state-based observers and probabilistic modeling are potential approaches to resolve these discrepancies~\cite{hafner_mastering_2025, jain_smooth_2025}.

\clearpage\newpage
\section*{Acknowledgments}
TH is supported by the U.S. National Science Foundation Graduate Research Fellowship Program under Grant No. DGE 2140004. HR is supported by the U.S. National Science Foundation under Grant No. 2313998. Any opinions, findings, and conclusions or recommendations expressed in this material are those of the author(s) and do not necessarily reflect the views of the U.S. National Science Foundation. We thank Jesse Zhang for insightful discussions about experimentation.

\bibliography{references}

\clearpage
\newpage
\onecolumn
\appendix
\crefname{appendix}{Appendix}{Appendices}
\Crefname{appendix}{Appendix}{Appendices}
\crefalias{section}{appendix}
\crefalias{subsection}{appendix}
\crefalias{subsubsection}{appendix}
\clearpage
\onecolumn

\section*{Appendix Contents}
\vspace{4pt}
\noindent\rule{\linewidth}{0.8pt}
\begingroup
\newcommand{\ifappendix}[1]{%
  \IfBeginWith{\etocthenumber}{A}{#1}{%
    \IfBeginWith{\etocthenumber}{B}{#1}{%
      \IfBeginWith{\etocthenumber}{C}{#1}{%
        \IfBeginWith{\etocthenumber}{D}{#1}{%
          \IfBeginWith{\etocthenumber}{E}{#1}{%
            \IfBeginWith{\etocthenumber}{F}{#1}{%
              \IfBeginWith{\etocthenumber}{G}{#1}{%
                \IfBeginWith{\etocthenumber}{H}{#1}{%
                  \IfBeginWith{\etocthenumber}{I}{#1}{%
                    \IfBeginWith{\etocthenumber}{J}{#1}{}}}}}}}}}}%
                    }
                    \etocsetstyle{section}
                      {}
                        {\vspace{4pt}}
                          {\ifappendix{\noindent\textbf{\etocnumber}\hspace{1em}%
                              \textbf{\etocname}\dotfill\textbf{\etocpage}\par}}
                                {}
                                \etocsetstyle{subsection}
                                  {\vspace{2pt}}
                                    {}
                                      {\ifappendix{\noindent\hspace{1em}\etocnumber\hspace{1em}%
                                          \etocname\dotfill\etocpage\par}}
                                            {}
                                            \etocsettocstyle{}{}
                                            \etocsetnexttocdepth{subsection}
                                            \tableofcontents
                                            \endgroup

\vspace{6pt}
\noindent\rule{\linewidth}{0.8pt}

\clearpage

\clearpage

\section*{Website}
Experiment videos, code, and more can be found at:\\ \texttt{\website}

\section{Algorithm \& Architecture}\label{app:alg-arch}

The following section contains details describing the \name algorithm, split into training and planning (acting). Architectural diagrams are included in \Cref{app:architecture} to provide visual aid.

\subsection{\name Algorithms}\label{app:algorithms}
\subsubsection{Training}\label{app:alg-mpail2}

\begin{algorithm*}[h!]
\caption{\name}
\label{alg:mpail2}

\begin{algorithmic}[1]
\Require $\expd \subset \mathcal{O} \times \mathcal{O}$ \Comment{Task Observations}
\RequireCont $\enc:\mathcal{O}\rightarrow \mathcal{Z}$ \Comment{Encoder}
\RequireCont $\dyn:\zspc\times\mathcal{A}\rightarrow \zspc$ \Comment{Dynamics}
\RequireCont $\rew:\zspc\times\zspc\rightarrow \R$ \Comment{Inferred Reward}
\RequireCont $\val:\zspc\times\aspc\rightarrow\R$ \Comment{Value}
\RequireCont $\seq{a}_t\sim\pol(\cdot|z_t)$ \Comment{Policy}
\RequireCont $\seq{a}_t\sim\widehat{\Pi}(\cdot|\seq{a},z_t\,;\,\dyn,\rew,\val,\pol)$ \Comment{Planner}
\RequireCont $\buf \coloneqq \left\{\right\}$ \Comment{Replay Buffer}

\While{learning}
    \State Interact using planner (\Cref{alg:mppi-procedure})
    \begin{gather}
        \buf\gets\buf\cup \{(o_t,a_t,o_{t+1})\}_{t=1}^{T}
    \end{gather}
    \For{updates per episode}
    \State Sample trajectories and task observations
    \begin{gather}
        \{(o_t,a_t,o_{t+1})\}_{t=1}^{H}\sim\buf\,,\,
        (o,o')\in \expd \label{eq:alg-sample}
    \end{gather}
    \State Update Encoder and Dynamics  \eqref{eq:enc-dyn-loss}
    \begin{gather}
        \mathcal{L}_{e,f}(\omega,\psi) = \mathbb{E}_{\buf}\left[ \sum^{t+H}_{t'=t} \rho^{t'-t} \left\lVert \hat{z}_{t'} - \mathrm{sg}\left(z_{t'}\right) \right\rVert^2_2 \right]\label{eq:alg-dyn-jep}
    \end{gather}
    \State Update Inferred Reward
    \begin{gather}
        \mathcal{L}_{r}(\theta) = \mathbb{E}_{\enc(\buf)}[r] - \mathbb{E}_{\enc(\expd)}[r] + \beta \, \mathrm{GP}\left(r, \buf, \expd\right)\label{eq:alg-rew}
    \end{gather}
    \State Update Value
    \begin{gather}
        \mathcal{L}_Q(\zeta) = \mathbb{E}_{(z,a,z')\sim\enc(\buf),\,\hat{\seq{a}}'\sim\pol(\cdot|z')}\left[ \left(\val(z,a) - \left[r_\theta(z,z') + \gamma\,\bar{G}^\lambda(z',\hat{\seq{a}}')\right]\right)^2\right]\label{eq:alg-val}
    \end{gather}
    \State Update Policy \eqref{eq:policy-obj}
    \begin{equation}
        \mathcal{L}_\pi(\phi) = -\mathbb{E}_{\hat{\seq{a}}\sim\pol(\cdot|z)}\left[ G^\lambda(z,\hat{\seq{a}}) \right]
    \end{equation}
\EndFor
\EndWhile
\end{algorithmic}
\end{algorithm*}

\subsubsection{Gradient Penalty}\label{app:wasserstein-gp}

\begin{equation} \mathrm{GP}(r, \mathcal{B}, \mathcal{D}) = \mathbb{E}_{(\tilde{z}, \tilde{z}')}\!\left[ \left(\left\|\nabla_{(\tilde{z}, \tilde{z}')} r(\tilde{z}, \tilde{z}')\right\|_2 - 1\right)^2 \right] , \label{eq:gp} \end{equation}
with $(\tilde{z}, \tilde{z}')$ uniform linear interpolations between
expert latents $e_\omega(\mathcal{D})$ and buffer latents $e_\omega(\mathcal{B})$~\cite{gulrajani_improved_2017}. 

\subsubsection{Planning}\label{app:planning}

\clearpage
\begin{algorithm*}[h!]
\caption{\textsc{\name (Planning)}}\label{alg:mppi-procedure}
\begin{algorithmic}[1]
\Require 
\Statex Number of trajectories to sample $N$; 
\Statex Proportion of trajectories sampled from policy $M/N$;
\Statex Number of elite samples $K$
\Statex Planning horizon $H$;
\Statex Number of optimization iterations $J$
\Statex Sampling standard deviation (std) clip range $(\sigma_{\textrm{min}}, \sigma_{\textrm{max}})$
\Statex Softmax temperature $\eta$
\Statex \addmpail{$\enc:\mathcal{O}\rightarrow \mathcal{Z}$} \Comment{Encoder}
\Statex \addmpail{$\dyn:\zspc\times\mathbb{R}^m\times\mathcal{A}\rightarrow \zspc\times\mathbb{R}^m$} \Comment{Latent Dynamics}
\Statex \addmpail{$\rew:\zspc\times\zspc\rightarrow \R$} \Comment{Learned Reward}
\Statex \addmpail{$\val:\zspc\times\aspc\rightarrow\R$} \Comment{Value}
\Statex \addmpail{$\seq{a}\sim\pol(\cdot|z)$} \Comment{Multi-step Policy}
\vspace{0.5em}
\State \textbf{Procedure} \textsc{Plan}$(o_t, \mathbf{a}_{t-1})$
\vspace{0.5em}
\State $z_t\gets \addmpail{\enc(o_t)}$
\State $(\boldsymbol{\mu}_{t}^0)_i \gets (\mathbf{a}_{t-1})_{i+1}$ \hfill $\triangleright$ \texttt{Roll previous plan one timestep backward for next sampling mean}
\State $(\boldsymbol{\mu}_{t}^0)_H \gets 0$\hfill $\triangleright$ \texttt{Set sampling mean to 0 for last timestep}
\State $\Sigma_{ii}^0 \gets \sigma_{\textrm{max}}$\hfill $\triangleright$ \texttt{Set sampling std to max for first iteration}
\State $\Sigma_{ab,a\neq b} \gets 0$\hfill $\triangleright$ \texttt{Isotropic sampling}
\For{$j \gets 0$ \textbf{to} $J-1$}
    \For{$k \gets 0$ \textbf{to} $N - 1$} \hfill  $\triangleright$ \texttt{Model rollouts and return estimation (parallelized)}
        \State $\hat{z}^k_t \gets z_t$
        \If{$k < N - M$}
        \State $\mathbf{a}_t^k \sim \mathcal{N}(\boldsymbol{\mu}_{t}^j, \Sigma^j)$ \hfill  $\triangleright$ \texttt{Sample random plan}
        \Else 
        \State $\mathbf{a}_t^k \sim \addmpail{\pol(\mathbf{a}|z_t)}$ \hfill $\triangleright$ \texttt{Sample plan from policy}
        \EndIf
        \For{$t' \gets t$ \textbf{to} $t+H-1$}
            \State $\hat{z}^k_{t'+1} \gets \addmpail{\dyn(\hat{z}^k_{t'}, a^k_{t'})}$  \hfill  $\triangleright$ \texttt{Predict next latent state}
            \State $\hat{r}_{t'}^k \gets$~\addmpail{$\rew(\hat{z}_{t'}^k, \hat{z}_{t'+1}^k)$} \hfill  $\triangleright$ \texttt{Compute latent state-transition rewards}
        \EndFor
    \State $\mathcal{R}(\tau_k) \gets$ $\gamma^H$\addmpail{$\val(\hat{z}_{t+H}^k,a^k_{t+H})$} + $\sum_{t'=0}^{H-1}\gamma^{t'}\hat{r}_{t'}^k$ \hfill  $\triangleright$ \texttt{Total trajectory return}
    \EndFor
    \State $\mathcal{E} \gets$ indices of the $K$ largest $\mathcal{R}(\tau_k)$ \hfill $\triangleright$ \texttt{Elite set}
    \State $\beta \gets \max_{k\in\mathcal{E}}\mathcal{R}(\tau_k)$
    \State $\mathcal{Z} \gets \sum_{k\in\mathcal{E}}\exp\!\left(\tfrac{1}{\eta}\left(\mathcal{R}(\tau_k)-\beta\right)\right)$\hfill  $\triangleright$ \texttt{Non-elites are excluded from the normalization}
    \For{$k \in \mathcal{E}$}
        \State $w(\tau_k) \gets \frac{1}{\mathcal{Z}}\exp\!\left(\tfrac{1}{\eta}\left(\mathcal{R}(\tau_k) - \beta\right)\right)$ \hfill  $\triangleright$ \texttt{Weights are a trajectory's score over the total}
    \EndFor
    \State $\boldsymbol{\mu}_{t}^j \gets \sum_{k\in\mathcal{E}}w(\tau_k)\,\mathbf{a}^k_t$
    \State $\boldsymbol{\sigma}_{t}^j \gets \sqrt{\sum_{k\in\mathcal{E}}w(\tau_k)\left(\mathbf{a}^k_t - \boldsymbol{\mu}_{t}^j\right)^2}$
    \State $\Sigma_{ii}^j \gets \operatorname{clip}(\boldsymbol{\sigma}^j,\sigma_{\textrm{min}}, \sigma_{\textrm{max}})$
\EndFor
\State $\mathbf{a}_t \sim \mathcal{N}(\boldsymbol{\mu}_{t}^J, \Sigma^J)$ \hfill $\triangleright$ \texttt{Sample new plan from optimized distribution}
\State\Return $\mathbf{a}_t$
\vspace{0.5em}
\State \textbf{End Procedure}
\end{algorithmic}
\end{algorithm*}

The planning algorithm is shown in~\Cref{alg:mppi-procedure}. It largely matches the implementation of MPPI~\cite{williams_information-theoretic_2018, hansen_td-mpc2_2023} with the learned dynamics, reward, and value models dropped in (highlighted in blue in~\Cref{alg:mppi-procedure}). Only the $K$ elites enter the softmax, so the planner is invariant to additive shifts of the reward and value. Its objective can be written,
\begin{equation}
\argmax_{\boldsymbol{\mu}_t,\boldsymbol{\sigma}_t} \mathbb{E}_{\mathbf{a}_t\sim\mathcal{N}(\boldsymbol{\mu}_t,\boldsymbol{\sigma}_t)}\left[ \gamma^H\hat q_{t+H} + \sum^{H-1}_{t'=t}  \gamma^{t'-t}\hat r_{t'}\right]\label{eq:mppi}
\end{equation}
where the predicted rewards $\hat{r}$ and values $\hat{q}$ are used to estimate the return of each plan $\mathbf{a}_t$, rolled out as in \Cref{sec:method}.

\subsection{Related Designs: Latent Planning and Action Chunking}\label{app:related-designs}

Planning-based learners are efficient in simulation~\cite{hansen_td-mpc2_2023, jain_smooth_2025} and robust on real robots~\cite{han_model_2025}. TD-MPC2~\cite{hansen_td-mpc2_2023} trains its latent with reward and value losses, which presupposes a known, stationary reward and ties the dynamics to the task. \name's encoder and dynamics train only on $(o,a,o')$, before any task or demonstration is defined, closer to joint-embedding predictive architectures~\cite{lecun_path_nodate, grill_bootstrap_2020}. Under an inferred reward, the latent becomes shaped by a discriminator, introducing adversarial representations which can substantially hurt prediction (see experiments in~\Cref{app:tdmpc2-latent}). \name's multi-step policy is also similar but differs crucially from action chunking in practice~\cite{zhao_learning_2023}: chunking in model-free paradigms necessarily expand the critic dimensionality by the horizon, whereas \name unrolls the action sequence through the dynamics, improving the online sample efficiency of the critic while retaining the multi-step policy.

\subsection{\name Architecture}\label{app:architecture}

\begin{figure}[h]
    \centering
\begin{subfigure}[h]{0.49\textwidth}
    \centering
    \includegraphics[width=\textwidth]{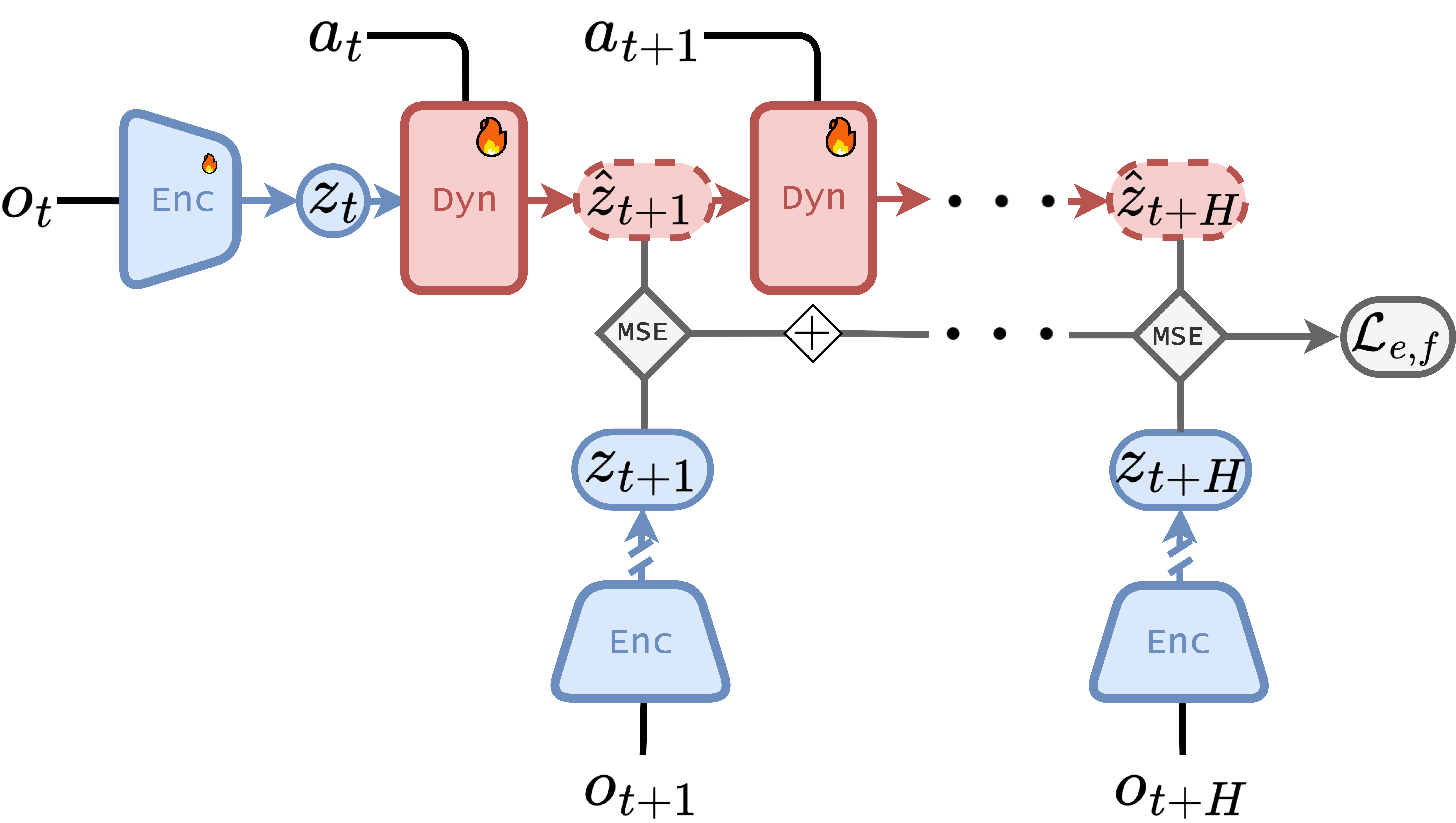}
    \caption{\textbf{Encoder and Dynamics Training.} The encoder $\enc$ maps observation $o_t$ to latent state $z_t$. Dynamics $\dyn$ auto-regressively predicts future latent states $\hat{z}_{t+1}, \ldots, \hat{z}_{t+H}$ given action inputs. Training minimizes the multi-step MSE loss $\mathcal{L}_{e,f}$ against stop-gradient targets obtained by fresh encoder passes on subsequent observations (Eq.~\ref{eq:alg-dyn-jep}).}
    \label{fig:arch-dyn}
\end{subfigure}
\hfill
\begin{subfigure}[h]{0.49\textwidth}
    \centering
    \includegraphics[width=\textwidth]{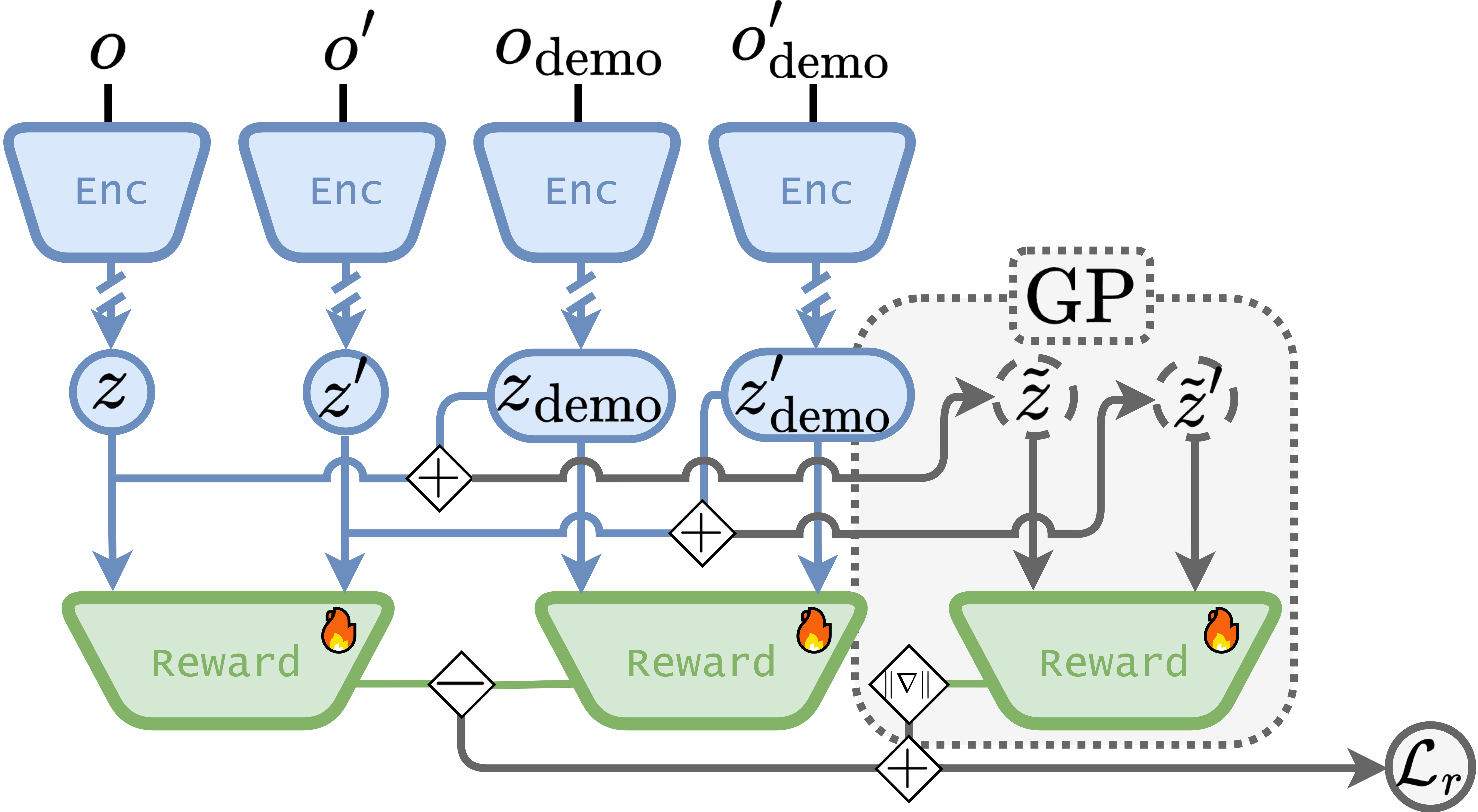}
    \caption{\textbf{Reward Training.} The reward model $\rew$ is trained contrastively: agent transition scores $(z, z')$ are minimized while demonstration transition scores $(z_{\text{demo}}, z'_{\text{demo}})$ are maximized. A gradient penalty (GP) regularizes the reward function over interpolated latent pairs $(\tilde{z}, \tilde{z}')$, yielding loss $\mathcal{L}_r$ (Eq.~\ref{eq:alg-rew}).}
    \label{fig:arch-rew}
\end{subfigure}
\end{figure}

\begin{figure*}[h]
    \centering
    \includegraphics[width=.80\textwidth]{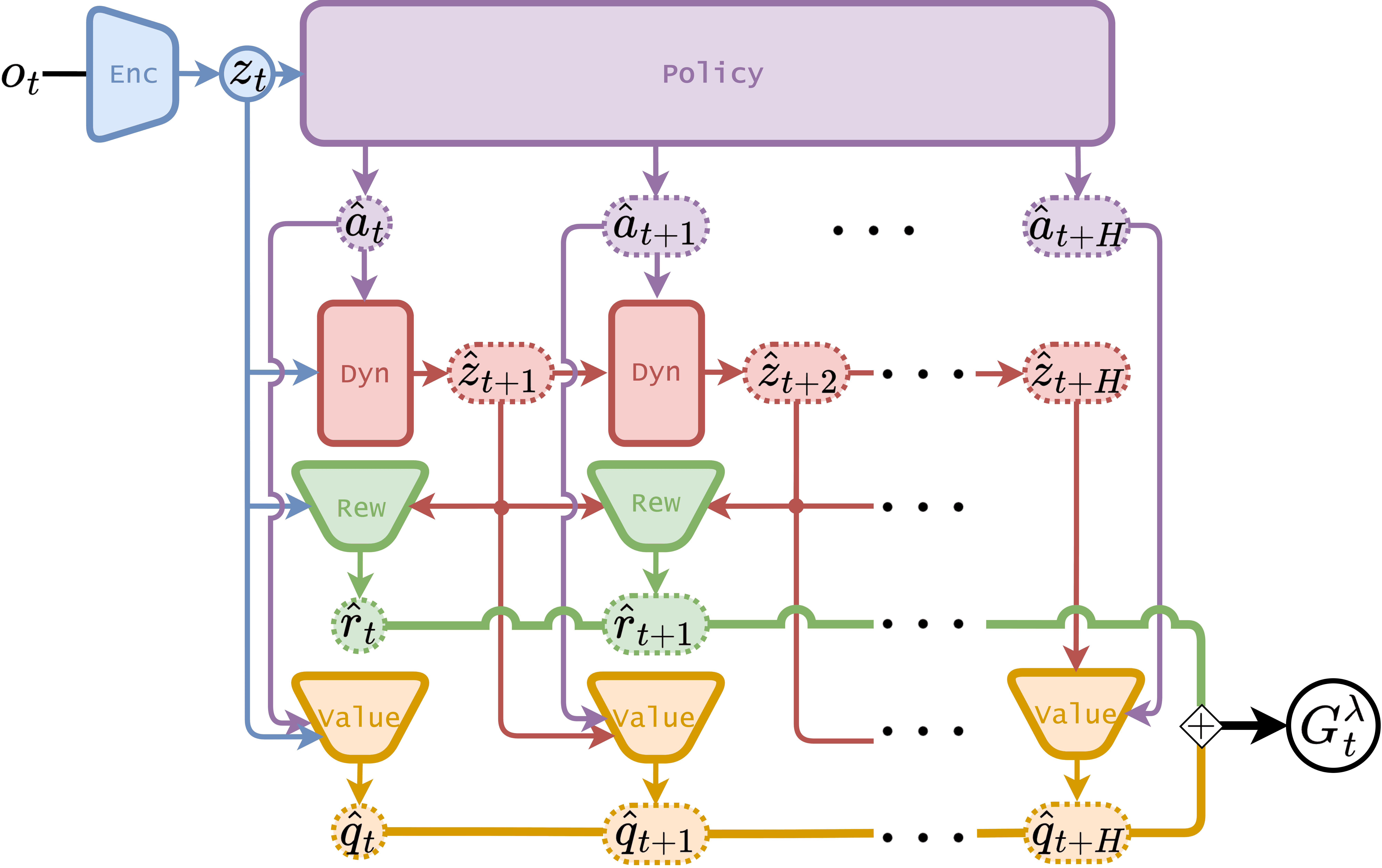}
    \caption{\textbf{\name $\lambda$-Return.} Components are unrolled over horizon $H$. The encoder $\enc$ produces $z_t$; the policy produces action sequences $\hat{a}_{t}, \ldots, \hat{a}_{t+H}$; dynamics $\dyn$ predicts next latent states; reward $\rew$ scores transitions; value $Q$ estimates returns. With the correct coefficients, these predictions compute the $\lambda$-return $G_t^\lambda$ for actor-critic updates (\Cref{eq:lam-return})}
    \label{fig:arch-full}
\end{figure*}

\Cref{fig:arch-dyn,fig:arch-rew,fig:arch-full} illustrate the encoder/dynamics training objective, the reward training objective, and the full \name architecture respectively. The PyTorch module summary can be found in \Cref{fig:pytorch}.

\begin{figure}[t]
\centering
\begin{Verbatim}[fontsize=\tiny,baselinestretch=0.8]
[INFO] Planner initialized. Total number of params: 7114494
[INFO]  Encoder: 2327744
[INFO]  Dynamics: 793088
[INFO]  Reward: 787968
[INFO]  Value: 2649605
[INFO]  Sampling: 556088
Planner(
    (encoder): MultiCoder(
        (coders): ModuleList(
            (0): Coder(
                (coder): Sequential(
                    (0): Linear(in_features=8, out_features=256, bias=True)
                    (1): SiLU()
                    (2): Linear(in_features=256, out_features=512, bias=True)
                    (3): LayerNorm((512,), eps=1e-05, elementwise_affine=True)
                    (4): SiLU()
                )
            )
            (1-2): 2 x CNNCoder(
                (coder): Sequential(
                    (0): Conv2d(3, 32, kernel_size=(3, 3), stride=(2, 2))
                    (1): SiLU()
                    (2): Conv2d(32, 32, kernel_size=(3, 3), stride=(2, 2))
                    (3): SiLU()
                    (4): Conv2d(32, 32, kernel_size=(3, 3), stride=(2, 2))
                    (5): SiLU()
                    (6): Conv2d(32, 32, kernel_size=(3, 3), stride=(1, 1))
                    (7): SiLU()
                    (8): Flatten(start_dim=1, end_dim=-1)
                    (9): Linear(in_features=800, out_features=512, bias=True)
                    (10): SiLU()
                )
            )
        )
        (latent_coder): Sequential(
            (0): Linear(in_features=1536, out_features=512, bias=True)
            (1): LayerNorm((512,), eps=1e-05, elementwise_affine=True)
            (2): SiLU()
            (3): Linear(in_features=512, out_features=512, bias=True)
            (4): LayerNorm((512,), eps=1e-05, elementwise_affine=True)
            (5): SiLU()
            (6): Linear(in_features=512, out_features=512, bias=True)
            (7): LayerNorm((512,), eps=1e-05, elementwise_affine=True)
        )
    )
    (dynamics): Dynamics(
        (model): Sequential(
            (0): Linear(in_features=516, out_features=512, bias=True)
            (1): LayerNorm((512,), eps=1e-05, elementwise_affine=True)
            (2): SiLU()
            (3): Linear(in_features=512, out_features=512, bias=True)
            (4): LayerNorm((512,), eps=1e-05, elementwise_affine=True)
            (5): SiLU()
            (6): Linear(in_features=512, out_features=512, bias=True)
            (7): LayerNorm((512,), eps=1e-05, elementwise_affine=True)
        )
    )
    (reward): Reward(
        (model): Sequential(
            (0): Linear(in_features=1024, out_features=512, bias=True)
            (1): SiLU()
            (2): Linear(in_features=512, out_features=512, bias=True)
            (3): SiLU()
            (4): Linear(in_features=512, out_features=1, bias=False)
        )
    )
    (value): EnsembleValue(
        (Qs): ModuleList(
            (0-1): 2 x Q(
                (model): Sequential(
                    (0): Linear(in_features=516, out_features=512, bias=True)
                    (1): LayerNorm((512,), eps=1e-05, elementwise_affine=True)
                    (2): SiLU()
                    (3): Linear(in_features=512, out_features=512, bias=True)
                    (4): LayerNorm((512,), eps=1e-05, elementwise_affine=True)
                    (5): SiLU()
                    (6): Linear(in_features=512, out_features=1, bias=True)
                )
            )
        )
    )
    (sampling): PolicySampling(
        (policy): PolicyNetwork(
            (model): Sequential(
                (0): Linear(in_features=512, out_features=512, bias=True)
                (1): LayerNorm((512,), eps=1e-05, elementwise_affine=True)
                (2): SiLU()
                (3): Linear(in_features=512, out_features=512, bias=True)
                (4): LayerNorm((512,), eps=1e-05, elementwise_affine=True)
                (5): SiLU()
                (6): Linear(in_features=512, out_features=56, bias=True)
            )
        )
    )
)
\end{Verbatim}
\caption{\textbf{PyTorch~\citep{paszke_pytorch_2019} 7.1M Model Architecture} for observation space with two RGB images and proprioception (i.e. Push Block and Pick-and-Place).}\label{fig:pytorch}
\end{figure}

\section{Experiment Setup}\label{app:exp-details}

We perform experiments in visual manipulation for two main reasons: (1) Planning-based IRL has been demonstrated to perform well in navigation~\cite{han_model_2025} but it is not yet clear whether the planning benefits go beyond self-prediction towards real-world dynamics, like object interactions. (2) Simulation, rewards, and action supervision are all practical bottlenecks in visual manipulation, making it a challenging yet realistic setting for IRLfO. Indeed, our baselines in RL and BC have been widely validated in visual manipulation~\cite{ball_efficient_2023,chi_diffusion_2025}. Best checkpoints are selected by evaluating all automatically saved checkpoints (every 10 episodes) beyond where the learner generally begins to exhibit successes (e.g. after 120 episodes in Mug-on-Plate).

\mypara{Resets.} Reset locations are sampled programmatically rather than chosen by the operator during training; objects are reset by hand between episodes, no mid-episode intervention occurs, and no run required recovery from a hardware or tracking failure.

\begin{table}[h]
\centering
\footnotesize
\setlength{\tabcolsep}{4pt}
\scalebox{0.8}{
\begin{tabular}{l l l r}
\toprule
\textbf{Task} & \textbf{Demo Observations} & \textbf{Object dim. (reset region)} & \textbf{\# of Demos} \\
\midrule
\rowcolor{gray!20}\multicolumn{4}{l}{\textit{Simulation --- IsaacLab, Franka arm; 5 seeds}} \\
Block Push (state) & state & $4{\times}4{\times}4$\,cm ($10{\times}6$\,cm) & 27 \\
Block Push         & images + proprioception & $4{\times}4{\times}4$\,cm ($10{\times}6$\,cm) & 27 \\
Pick-and-Place     & images + proprioception & $4{\times}4{\times}4$\,cm ($10{\times}6$\,cm) & 30 \\
\midrule
\rowcolor{gray!20}\multicolumn{4}{l}{\textit{Real --- $64{\times}64$ RGB (external\,+\,wrist); Franka: Push, Kinova Gen3: PnP; 3 seeds}} \\
Block Push         & images + proprioception & $5{\times}5{\times}5$\,cm ($18{\times}18$\,cm)        & 10 \,(1{,}451 trans.) \\
Pick-and-Place     & images + proprioception & $4{\times}4{\times}3$\,cm ($8{\times}16$\,cm) & 10 \,(1{,}025 trans.) \\
Mug-on-Plate & images + proprioception & mug\,$\varnothing$8${\times}$9\,cm\,($18{\times}18$\,cm), plate\,$\varnothing$14\,cm\,($20{\times}30$\,cm) & 15\,(1{,}802 trans.) \\
Block Push (Video)    & external image only & $5{\times}5{\times}5$\,cm ($18{\times}18$\,cm) & 10 (914 trans.) \\
\bottomrule
\end{tabular}
}
\vspace{4pt}
\caption{\textbf{Experiment Setup Summary.} \textit{Block Push}: push cube past a target line. \textit{Pick-and-Place}: grasp, lift $15$\,cm, place past a target line.
Full details in~\Cref{app:push-setup,app:pnp-setup}.}
\label{tab:tasks}
\end{table}

\subsection{Checkpoint Evaluation Results}\label{app:ckpt-eval}

Full evaluation results are reported in \Cref{tab:sim-eval} (simulation) and in \Cref{tab:real-eval-summary} (real-world).

\begin{table}[h]
    \centering
    \scriptsize 
    \setlength{\tabcolsep}{0.70pt}
    \sisetup{
        separate-uncertainty,
        table-format=2.1(2.1), 
        detect-weight=true,
        tight-spacing=true
    }
    \begin{tabular*}{\columnwidth}{@{\extracolsep{\fill}} c SSSSSSS @{}}
        \toprule
        \textbf{Ckpt} & {\textbf{\name}} & {\textbf{\name}} & {\textbf{\name}} & {\textbf{\name}} & {\textbf{\name}} & {\textbf{RLPD}} & {\textbf{BC}} \\
        {\tiny (\# steps)} & & {\textbf {[-P] (MAIRL)}} & {\textbf{[-PM] (DAC)}} & {\textbf{[-PMO] (AIRL)}} & {\textbf{[-O] (MPAIL)}} & & {\textbf{(Diffusion)}} \\
        \midrule
        \multicolumn{8}{l}{\cellcolor{gray!15}\textit{Sim: Block Push (State)}} \\
        100 & 64.0 \pm 13.0 & 21.0 \pm 14.7 & 14.0 \pm 14.0 & 0.0 \pm 0.0 & 0.0          & 4.0 \pm 5.9   & 60.2 \pm 1.6 \\
        200 & 76.0 \pm 8.1  & 6.0 \pm 6.0   & 43.0 \pm 21.0 & 0.0 \pm 0.0 & 0.0          & 35.4 \pm 4.1  & 60.2 \pm 1.6 \\
        300 & 71.0 \pm 13.7 & 4.0 \pm 4.0   & 36.0 \pm 22.1 & 8.0 \pm 8.0 & 4.0 \pm 8.9  & 54.8 \pm 11.5 & 60.2 \pm 1.6 \\
        400 & 87.0 \pm 6.4  & 0.0 \pm 0.0   & 24.0 \pm 19.4 & 0.0 \pm 0.0 & 22.4 \pm 32.9& 58.2 \pm 9.0  & 60.2 \pm 1.6 \\
        500 & 88.0 \pm 6.0  & 32.0 \pm 19.4 & 25.0 \pm 10.7 & 8.0 \pm 8.0 & 35.8 \pm 39.0& 60.4 \pm 12.5 & 60.2 \pm 1.6 \\
        \midrule
        \multicolumn{8}{l}{\cellcolor{gray!15}\textit{Sim: Block Push (Image)}} \\
        100 & 78.0 \pm 7.0  & 40.0 \pm 24.5 & 0.0 \pm 0.0   & 0.0 \pm 0.0   & 0.0         & 0.0 \pm 0.0 & 31.6 \pm 2.0 \\
        200 & 80.0 \pm 7.6  & 49.0 \pm 21.8 & 15.0 \pm 15.0 & 0.0 \pm 0.0   & 0.0         & 0.0 \pm 0.0 & 31.6 \pm 2.0 \\
        300 & 82.0 \pm 5.1  & 0.0 \pm 0.0   & 19.0 \pm 16.6 & 16.0 \pm 16.0 & 3.8 \pm 8.5 & 0.0 \pm 0.0 & 31.6 \pm 2.0 \\
        400 & 76.0 \pm 7.8  & 9.0 \pm 9.0   & 32.0 \pm 19.8 & 0.0 \pm 0.0   & 0.0         & 0.0 \pm 0.0 & 31.6 \pm 2.0 \\
        500 & 73.0 \pm 13.3 & 35.0 \pm 15.2 & 59.0 \pm 20.8 & 0.0 \pm 0.0   & 0.0         & 0.0 \pm 0.0 & 31.6 \pm 2.0 \\
        \midrule
        \multicolumn{8}{l}{\cellcolor{gray!15}\textit{Sim: Pick and Place (Image)}} \\
        100 & 9.0 \pm 9.0   & 13.0 \pm 7.0  & 0.0 \pm 0.0   & 0.0 \pm 0.0   & 0.0         & 0.0 \pm 0.0 & 46.6 \pm 2.3 \\
        200 & 33.0 \pm 15.0 & 17.0 \pm 17.0 & 0.0 \pm 0.0   & 0.0 \pm 0.0   & 0.0         & 0.0 \pm 0.0 & 46.6 \pm 2.3 \\
        300 & 58.0 \pm 14.5 & 32.0 \pm 20.2 & 2.0 \pm 2.0   & 0.0 \pm 0.0   & 0.0         & 0.0 \pm 0.0 & 46.6 \pm 2.3 \\
        400 & 58.0 \pm 13.7 & 29.0 \pm 18.5 & 27.0 \pm 14.6 & 0.0 \pm 0.0   & 0.0         & 0.0 \pm 0.0 & 46.6 \pm 2.3 \\
        500 & 50.0 \pm 8.4  & 41.0 \pm 21.4 & 33.0 \pm 19.6 & 0.0 \pm 0.0   & 0.0         & 0.0 \pm 0.0 & 46.6 \pm 2.3 \\
        \bottomrule
    \end{tabular*}
    \caption{\textbf{Full Checkpoint History by Environment Steps in Simulation.} BC (Diffusion) is offline; its performance is constant across interaction budgets.}
    \label{tab:sim-eval}
\end{table}

\begin{table}[h]
\centering
\scriptsize
\setlength{\tabcolsep}{4pt}
\begin{tabular*}{\columnwidth}{@{\extracolsep{\fill}} c ccccc @{}}
\toprule
\textbf{Ckpt} & \textbf{\name} & \textbf{[-P] (MAIRL)} & \textbf{[-PM] (DAC)} & \textbf{RLPD} & \textbf{BC (Diff.)} \\
\midrule
\multicolumn{6}{l}{\cellcolor{gray!15}\textit{Real: Block Push}} \\
Last      & $61.8 \pm 13.5$            & $34.0 \pm 17.7$            & $0.0 \pm 0.0$ & $0.0 \pm 0.0$ & $\mathbf{94.1 \pm 5.9}$ \\
Best (\#) & $\mathbf{100 \pm 0}$ (100)  & $64.0 \pm 17.7$ (130)     & $0.0 \pm 0.0$ (---) & $0.0 \pm 0.0$ (---) & $\mathbf{94.1 \pm 5.9}$ (---) \\
\midrule
\multicolumn{6}{l}{\cellcolor{gray!15}\textit{Real: Pick-and-Place}} \\
Last      & $\mathbf{68.0 \pm 13.6}$   & $16 \pm 14.8$            & $0.0 \pm 0.0$ & $0.0 \pm 0.0$ & $12.0 \pm 5.8$ \\
Best (\#) & $\mathbf{82.0 \pm 12.2}$ (140)   & $16.0 \pm 14.8$ (150)    & $0.0 \pm 0.0$ (---) & $0.0 \pm 0.0$ (---) & $12.0 \pm 5.8$ (---) \\
\midrule
\multicolumn{6}{l}{\cellcolor{gray!15}\textit{Real: Mug-on-Plate}} \\
Last      & $\mathbf{35.3 \pm 17.3}$ & $2.0 \pm 2.8$ & $0.0 \pm 0.0$ & $0.0 \pm 0.0$ & $\mathbf{33.5 \pm 12.3}$ \\
Best (\#) & $\mathbf{54.9 \pm 14.7}$ (150)   & $3.9 \pm 2.8$ (190)  & $0.0 \pm 0.0$ (---) & $0.0 \pm 0.0$ (---) & $33.5 \pm 12.3$ (---) \\
\midrule
\multicolumn{6}{l}{\cellcolor{gray!15}\textit{Real: Transfer Push (Transferred)}} \\
Last      & $\mathbf{63.5 \pm 4.1}$ & $8.1 \pm 5.7$ & $0.0 \pm 0.0$ & $0.0 \pm 0.0$ & $7.8 \pm 3.0$ \\
Best (\#) & $\mathbf{90.2 \pm 4.7}$ (60) & $13.9 \pm 13.3$ (130) & $0.0 \pm 0.0$ (---) & $0.0 \pm 0.0$ (---) & $15.8 \pm 7.8$ (---) \\
\midrule
\multicolumn{6}{l}{\cellcolor{gray!15}\textit{Real: Transfer Push (From Scratch)}} \\
Last      & $\mathbf{80.3 \pm 11.5}$ & $34.0 \pm 9.8$ & $0.0 \pm 0.0$ & $0.0 \pm 0.0$ & $26 \pm 9.8$ \\
Best (\#) & $\mathbf{88.2 \pm 5.7}$ (140) & $52.9 \pm 11.8$ (140) & $0.0 \pm 0.0$ (---) & $0.0 \pm 0.0$ (---) & $26 \pm 9.8$ (---) \\
\midrule
\multicolumn{6}{l}{\cellcolor{gray!15}\textit{Real: Transfer Pick-and-Place (Transferred)}} \\
Last      & $\mathbf{47 \pm 29.4}$ & $0.0 \pm 0.0$ & $0.0 \pm 0.0$ & $0.0 \pm 0.0$ & $14 \pm 3.3$ \\
Best (\#) & $\mathbf{94 \pm 10.2}$ (93) & $16 \pm 17.9$ (123) & $0.0 \pm 0.0$ (---) & $0.0 \pm 0.0$ (---) & $18 \pm 5.8$ (---) \\
\midrule
\multicolumn{6}{l}{\cellcolor{gray!15}\textit{Real: Transfer Pick-and-Place (From Scratch)}} \\
Last      & $\mathbf{43 \pm 29}$ & $6 \pm 5.8$ & $0.0 \pm 0.0$ & $0.0 \pm 0.0$ & $8 \pm 3.3$ \\
Best (\#) & $\mathbf{53 \pm 36.7}$ (143) & $12 \pm 11.7$ (153) & $0.0 \pm 0.0$ (---) & $0.0 \pm 0.0$ (---) & $12 \pm 5.8$ (---) \\
\multicolumn{6}{l}{\cellcolor{gray!15}\textit{Real: Block Push (Video-Only Demonstration)}} \\
Last      & $\mathbf{58.8 \pm 14.4}$ & $25.5 \pm 12.1$ & $0.0 \pm 0.0$ & $0.0 \pm 0.0$ & $23.5 \pm 8.3$ \\
Best (\#) & $\mathbf{62.7 \pm 10.0}$ (173) & $29.4 \pm 8.3$ (193) & $0.0 \pm 0.0$ (---) & $0.0 \pm 0.0$ (---) & $25.5 \pm 5.5$ (---) \\
\bottomrule
\end{tabular*}
\caption{\textbf{Real-World Results.} Success rate (\%) $\pm$ std across seeds by task across methods. Evaluations performed over 50 trials/task over 3 seeds. \textit{Last}: final-checkpoint success. \textit{Best (\#)}: best-checkpoint success with corresponding average environment-step count.}
\label{tab:real-eval-summary}
\end{table}

\subsection{Real-World Block Push Setup}\label{app:push-setup}
The real-world Block Push experiments are conducted on a Franka robotic arm. The observation space includes $64 \times 64$ RGB images from a fixed table-top RGB camera (Intel RealSense D435i), a wrist-mounted RGB camera (Intel RealSense D435i) rigidly attached to the arm's wrist and proprioception including joint position (7), joint velocities (7), end-effector Cartesian position (3) and a gripper state (1).
Actions are defined in the end-effector space as position velocity (3). All actions are bound to a clipped workspace ranging from [0.4, -0.25, 0.06] to [0.65, 0.25, 0.25].

For the task a $5\,\text{cm} \times 5\,\text{cm} \times 5\ \text{cm}$ cube is placed in a $18\,\text{cm} \times 18\ \text{cm}$ reset region. We collect a dataset of 10 demonstrations using a space-mouse totaling 1,451 transitions.

\textbf{Metrics.} Due to the high-quality instrumentation of the push setup, we are able to compute block positions via camera calibrations and AprilTags. One of the uses of these block positions is providing RLPD with dense reward.

Let $d_{\textrm{EE}}^{\textrm{B}}$ be the distance between the block and the center of the end-effector (gripper). Let $y_\textrm{goal}\coloneqq-0.1$ be the target $y$-threshold and let $y_B$ be the $y$-coordinate of the block. The dense reward is given by
\begin{equation}
    r(s) = -d_{\textrm{EE}}^{\textrm{B}} - \lvert y_\textrm{B} - y_\textrm{goal} \rvert.
\end{equation}

Success of an episode or evaluation is credited if any $y_B < y_\textrm{goal}$ at some point in the trajectory.

\subsection{Real-World Pick-and-Place Setup}\label{app:pnp-setup}
All real-world Pick-and-Place experiments are conducted on a Kinova Gen3 6-DoF robotic arm equipped with a Robotiq 2F-85 gripper. The observation space includes $64 \times 64$ RGB images from a fixed table-top RGB camera (Intel RealSense D435i), a wrist-mounted RGB camera (Intel RealSense D410) rigidly attached to the arm's wrist including joint position (6), joint velocities (6), end-effector Cartesian position (3) and a gripper state (2) - gripper state and gripper trigger.

At the beginning of each episode, a cuboid object ($4\,\text{cm} \times 4\,\text{cm} \times 3\ \text{cm}$, light blue) is randomly placed within an $8\,\text{cm} \times 16\ \text{cm}$ reset region. The minimum distance between the reset region and the target placement line is 18 cm.

Actions are defined in end-effector space and include Cartesian position commands in x,y, and z, together with a gripper command. All actions are constrained to a clipped workspace with bounds $[0.30, -0.15, 0.166]$ and $[0.47, 0.15, 0.27]$.

Demonstration data are collected in the real world via keyboard-based teleoperation, consisting of 10 demonstration episodes with a total of 1,025 transitions.

\textbf{Metrics.} The real-world pick-and-place setup does not have the capacity to provide dense rewards to RLPD. However, as done in~\cite{luo_serl_2025}, the operator is on standby to label progressions for reward classification. Within the limited interaction budget considered here, RLPD does not make contact with the block and so rarely receives reward, consistent with the difficulty of visual RL from scratch at this budget.

Success in the pick-and-place task is determined by completing four stages at any time in order: (i) initialized: $z_B<z_\textrm{min}$, (ii) lifted: $z_B\geq z_\textrm{min}$ and $y_B > y_\textrm{goal}$, (iii) transported: $z_B\geq z_\textrm{min}$ and $y_B \leq y_\textrm{goal}$, (iv) placed: $z_B\leq 0.03$ and $z_{\textrm{EE}} \leq 0.03$. These conditions are used in simulation to compute dense reward (using the Lagrangians of the conditions) for RLPD. In the real-world setup, stages are hand-labeled after training for evaluation metrics.

\subsection{Real-World Mug-on-Plate Setup}\label{app:mop-setup}
The real-world Mug-on-Plate experiments are conducted on a Franka robotic arm with the same observation space as the Block Push setup (\Cref{app:push-setup}). Actions are defined in end-effector space as position velocity with a gripper command~(4). All actions are bound to a clipped workspace ranging from $[0.4, -0.27, 0.085]$ to $[0.65, 0.27, 0.25]$.

At the beginning of each episode, a mug ($\varnothing 8\,\text{cm} \times 9$\,cm) is randomly placed within an $18\,\text{cm} \times 18\,\text{cm}$ reset region, and a plate ($\varnothing 14$\,cm) is randomly placed within a $20\,\text{cm}\times 30\,\text{cm}$ reset region. Demonstration data are collected via keyboard-based teleoperation, consisting of 15 episodes with a total of 1,802 transitions.

\textbf{Metrics.}
The mug-on-plate setup does not support automatic object tracking, so no dense reward is available for RLPD; as in the pick-and-place setup (\Cref{app:pnp-setup}), the operator labels reward during training via a keyboard interface. Success is defined as the mug resting stably on the plate at any point during the episode.

\subsection{Video-only Demonstration Setup}\label{app:exp-details-substate}

\begin{figure}[h]
    \centering
        \begin{minipage}[b]{.54\linewidth}
        \centering
        \includegraphics[width=\linewidth]{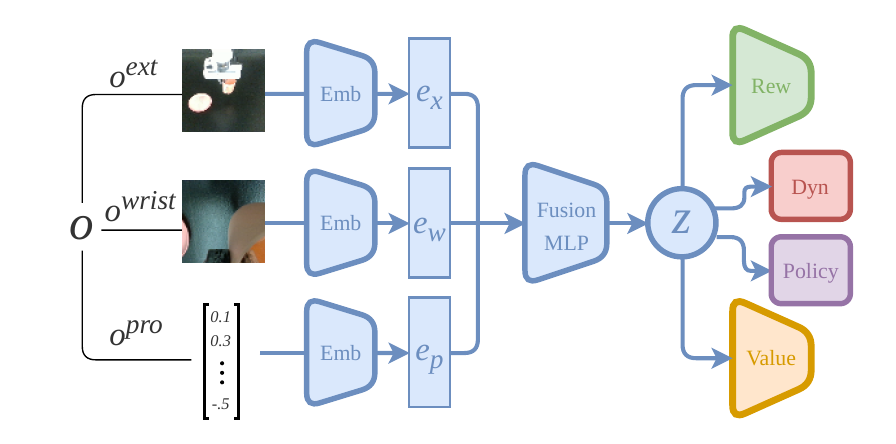}
        \caption{\textbf{Fused encoder} used in all experiments, except for video-only demonstration task. $o^\mathrm{ext}$ represents the \textit{external} video modality.}
        \label{fig:fuse-enc}
    \end{minipage}
    \hfill
    \begin{minipage}[b]{.45\linewidth}
        \centering
        \includegraphics[width=\linewidth]{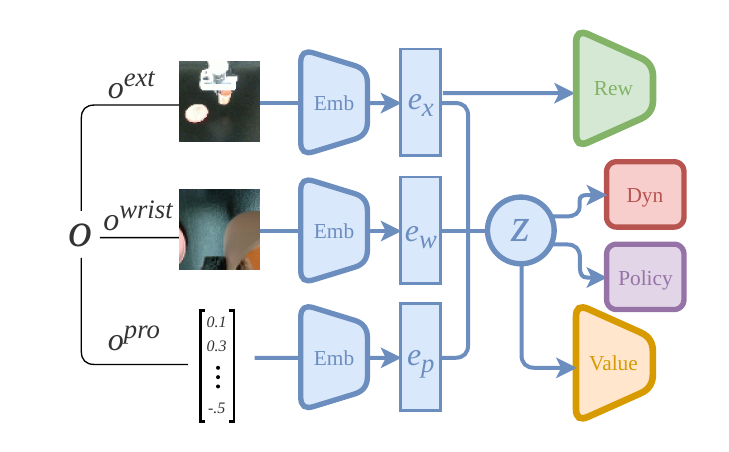}
        \caption{\textbf{Substate reward encoder} used in video-only demonstration task. $o^\mathrm{ext}$ represents the \textit{external} video modality.}
        \label{fig:sub-enc}
    \end{minipage}
\end{figure}

\Cref{fig:fuse-enc} and \Cref{fig:sub-enc} show the architectural modifications made for Block Push task from video-only demonstration. Inclusion of fusion MLP is used in most experiments mainly due to the popularity of the design~\cite{hansen_learning_2025, jain_smooth_2025}. However, our results reveal that it is equally possible to have the dynamics ``inherit'' the fusion process instead, keeping the latent representation compartmentalized. This enables rewarding a subset of the latent state towards scalable demonstration modalities, like third-person video. 

\clearpage
\section{Implementation}\label{app:imp-details}

With settings evaluated in this work, MPAIL2's action rate in ``Real:Push'' is 10.28 $\pm$ 0.45 Hz. Reported training times include online interaction, resets, and learner updates; they exclude checkpoint evaluations and, for RLPD, the operator's online reward labeling. In simulation without time-synchronization, the action rate is 22.05 $\pm$ 1.55 Hz. Parameters can be found in~\Cref{app:hyperparams}.

\subsection{Hyperparameters}\label{app:hyperparams}

\newcommand{\latentcolor}{NavyBlue}
\newcommand{\rewcolor}{ForestGreen}
\newcommand{\valcolor}{Orange}
\newcommand{\polcolor}{Violet}
\newcommand{\mppicolor}{Gray}

\newcommand{\colorcells}[3]{%
  \cellcolor{#1!10}\textcolor{#1}{#2} &
  \cellcolor{#1!10}\textcolor{#1}{#3}%
}
\newcommand{\latentcells}[2]{\colorcells{\latentcolor}{#1}{#2}}
\newcommand{\rewcells}[2]{\colorcells{\rewcolor}{#1}{#2}}
\newcommand{\valcells}[2]{\colorcells{\valcolor}{#1}{#2}}
\newcommand{\polcells}[2]{\colorcells{\polcolor}{#1}{#2}}
\newcommand{\mppicells}[2]{\colorcells{\mppicolor}{#1}{#2}}

\begin{table}[h]
    \centering
    \begin{tabular}{c|l c}
    \toprule
      Component & \textbf{Hyperparameter} &  \textbf{Value}\\
      \midrule
        & Optimizer & Adam \\
        & Learning Rate (LR) & 3e-4 \\
        & Return Parameter ($\lambda$) & 0.95 \\
        & Discount ($\gamma$) & 0.99 \\
        & Planning Horizon ($H$) & 7 \\
        & Replay Size & $\infty$ \\
        & UTD Ratio & 1.0 \\
        & Batch Size & 256 \\
        & Latent Dimension & 512 (state: 256) \\
        & Hidden Layers & [512, 512] \\
        & LayerNorm & True \\
      \midrule
        \multirow{2}{*}{\textbf{Encoder/Dynamics}}
        & \latentcells{Encoder LR}{3e-5} \\
        & \latentcells{Temporal Discount ($\rho$)}{0.95} \\
      \midrule
        \multirow{2}{*}{\textbf{Reward}}
        & \rewcells{GP Coefficient ($\beta$)}{0.1} \\
        & \rewcells{LayerNorm}{False} \\
      \midrule
        \multirow{3}{*}{\textbf{Value}}
        & \valcells{Polyak Coefficient}{0.01} \\
        & \valcells{Gradient Norm Clip}{5.0} \\
        & \valcells{Ensemble Size}{5} \\
      \midrule
        \multirow{3}{*}{\textbf{Policy}}
        & \polcells{Target Entropy}{$-|\mathcal{A}|$} \\
        & \polcells{Alpha LR}{3e-4} \\
        & \polcells{Gradient Norm Clip}{1.0} \\
      \midrule\midrule
        \multirow{6}{*}{\textbf{Planner}}
        & Temperature ($\eta$) & 2.0 \\
        & Number of Elites ($K$) & 64 \\
        & Iterations ($J$) & 5 \\
        & Policy Plan Fraction ($M/N$) & 0.05 \\
        & Number of Rollouts ($N$) & 512 \\
        & Std Range ($\sigma_{\min},\sigma_{\max}$) & (0.05, 2.0) \\
    \bottomrule
    \end{tabular}
    \caption{\textbf{MPAIL2 Hyperparameters}. Used across all experiments unless specified otherwise. Only the Reward model does not employ LayerNorm.}\label{tab:hps}
\end{table}

\subsection{Baseline Implementations}\label{app:baselines}

\begin{enumerate}
    \item \textbf{\namep (MAIRL~\cite{sun_adversarial_2021})} simply removes the online planning component from \name. When acting in the environment, \namep directly samples actions from a single-step policy $\pol(a_t|z_t)$. The policy is single-step as the multi-step policy is designed for reducing online planning computation. \namep is equivalent to \name in all other aspects: encoder, dynamics, reward, and value have equivalent losses. Where applicable, multi-step losses are computed through auto-regressive policy inference.
\item \textbf{\namepm (DAC~\cite{kostrikov_discriminator-actor-critic_2018})} removes both online planning and model-based components. Due to the large observation space and to reduce divergences from \name, \namepm improves on DAC by encoding observations to a latent space. Its encoder is updated via the value target loss and a scaled learning rate. \textit{We acknowledge that these modifications are not necessarily equivalent to \name, but we remark that various attempts were made at improving the model-free baseline and ensuring a fair ablation; for example, Random Fourier Features (RFF)~\cite{rahimi_random_2007} and no encoding did not exhibit noticeable improvements.}
    \item \textbf{\namepmo (AIRL~\cite{fu_learning_2018,orsini_what_2021})} takes the implementation of \namepm and modifies the replay buffer such that only interactions within the previous episode are able to be sampled.
\item \textbf{\nameo (MPAIL~\cite{han_model_2025}}) replaces the off-policy replay that \name uses for the reward, value-ensemble, and policy updates with an on-policy rollout
buffer, whereas the encoder and latent dynamics retain their off-policy JEP replay.
All other components, encoder, latent dynamics, WGAN-GP latent-transition
reward, five-head target-network Q-ensemble trained against
on-policy $\lambda$-return targets, and policy-seeded MPPI planner, are
unchanged from \name. Hyperparameters are reported in \Cref{tab:hps}.
We note that this ablation subsumes a faithful MPAIL implementation, which
underperformed in our preliminary experiments. To isolate the impact of
off-policy training rather than confound it with component-level deficits,
we retained the stronger MPAIL2 components listed above. We additionally
verified that replacing the five-head Q-ensemble with MPAIL's on-policy
$V_\phi$ value head, the value architecture that was specifically
designed to maximize on-policy sample efficiency, did not improve results. Instead, we report
\nameo{} as a strengthened MPAIL-style ablation that preserves the remaining \name{}
components.

\item \textbf{RLPD} is implemented as given in~\cite{luo_serl_2025} but is also improved via an encoder in \namepm. Its hyperparameters and architecture coincide with \namepm (\Cref{tab:dac-hps}); in place of \namepm's learned reward it uses the hand-designed reward or the online, human-labeled reward.
\item \textbf{BC} is implemented as a diffusion policy with a U-Net denoiser. Implementation and parameters are as given in~\cite{chi_diffusion_2025}. We train each policy for 500 epochs.
\end{enumerate}

Apart from BC, which follows its reference implementation~\cite{chi_diffusion_2025}, no component of any method (encoder, dynamics, reward, value, or policy) is pre-trained or frozen, and methods do not share encoders. All online methods learn only from their own interactions and the demonstrations of \Cref{tab:tasks}.

\begin{table}[h]
    \centering
    \begin{tabular}{l c}
    \toprule
      \textbf{Hyperparameter} & \textbf{Value}\\
      \midrule
        Learning Rate (reward, value, policy, $\alpha$) & 3e-4 \\
        Encoder Learning Rate (if applicable, value loss only) & 3e-5 \\
        Value Target & 1-step TD \\
        UTD Ratio & 1.0 \\
        Batch Size & 256 \\
        LayerNorm & True (except reward) \\
        Value Ensemble Size & 5 \\
        Reward GP Coefficient ($\beta$; \namepm only) & 0.1 \\
        Sampling Ratio (RLPD only) & 1:1 \\
    \bottomrule
    \end{tabular}
    \caption{\textbf{\namepm (DAC) and RLPD hyperparameters} (real-world configuration). RLPD has no learned reward, so the GP coefficient applies to \namepm only. Parameters not explicitly mentioned are shared with \Cref{tab:hps} where applicable.}\label{tab:dac-hps}
\end{table}

\clearpage
\section{Additional Results}\label{app:add-results}

\subsection{Task Overview}\label{app:task-overview}

\begin{figure*}[!htbp]
    \centering
    \begin{minipage}{0.30\textwidth}
        \centering
        \reflectbox{%
            \includegraphics[width=\linewidth, height=\linewidth, keepaspectratio=false]{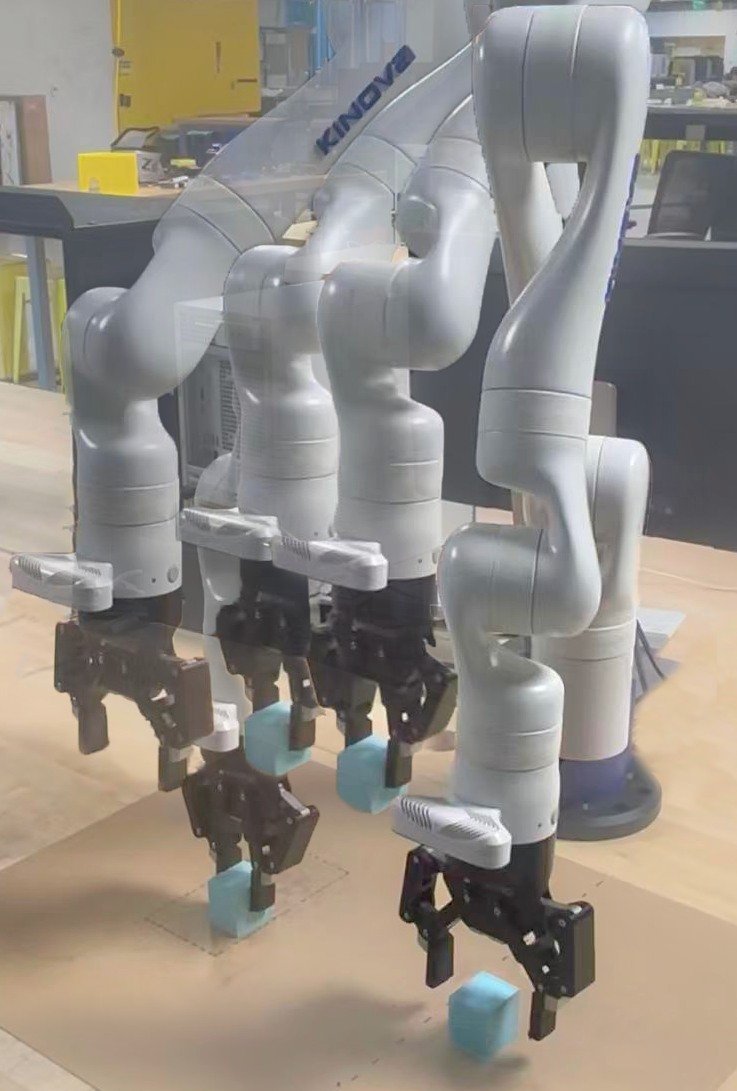}%
        }
        \vspace{-0.7cm}
        \par\medskip \small{(a) Pick and Place (Real)}
    \end{minipage}
    \hfill
    \begin{minipage}{0.30\textwidth}
        \centering
        \includegraphics[width=\linewidth, height=\linewidth, keepaspectratio=false]{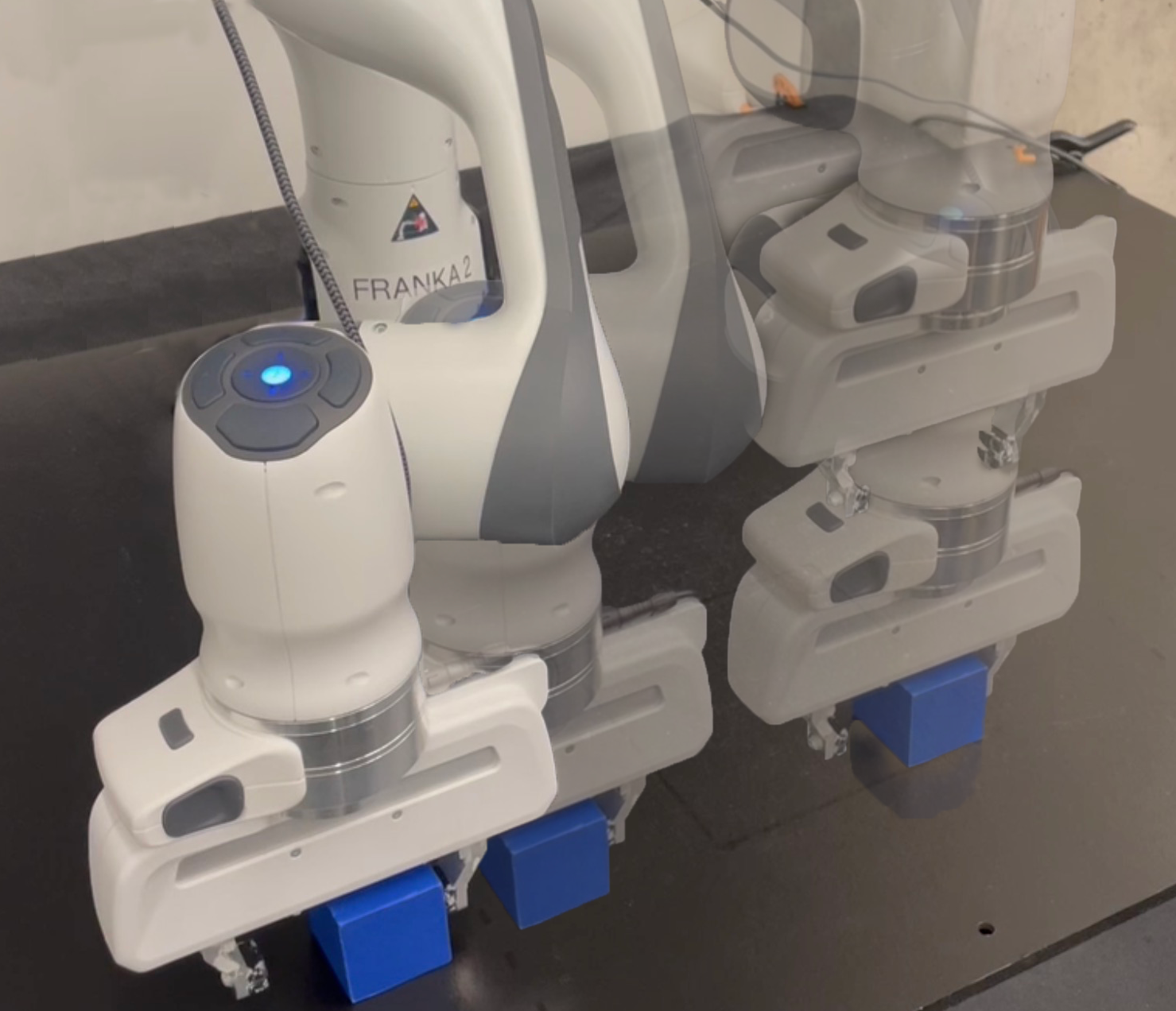}
        \vspace{-0.7cm}
        \par\medskip \small{(b) Push (Real)}
    \end{minipage}
    \hfill
    \begin{minipage}{0.30\textwidth}
        \centering
        \includegraphics[width=\linewidth, height=\linewidth, keepaspectratio=false]{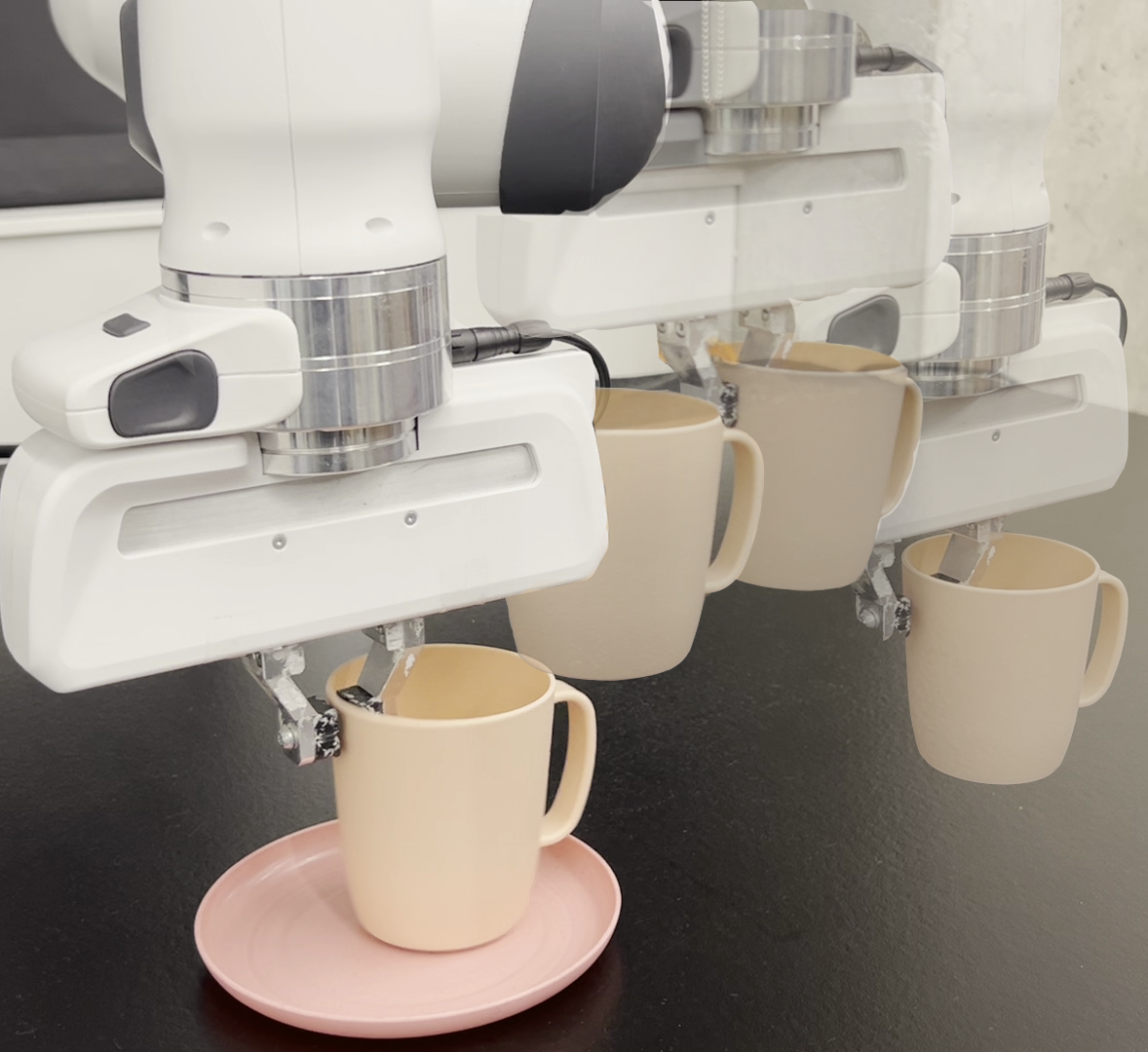}
        \vspace{-0.7cm}
        \par\medskip \small{(c) Mug-on-Plate (Real)}
    \end{minipage}
    \hfill
    \begin{minipage}{0.30\textwidth}
        \centering
        \includegraphics[width=\linewidth, height=\linewidth, keepaspectratio=false]{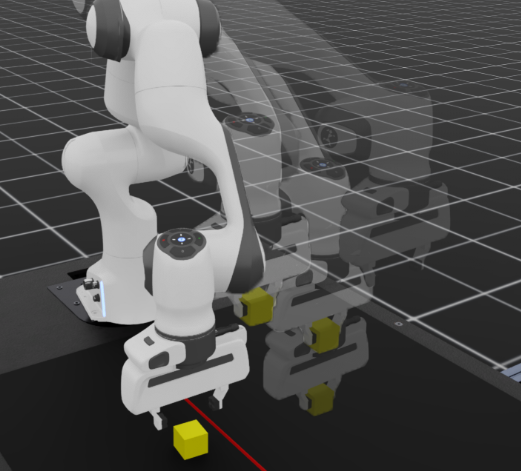}
        \vspace{-0.7cm}
        \par\medskip \small{(d) Pick and Place (Sim)}
    \end{minipage}
    \hspace{48pt}
    \begin{minipage}{0.30\textwidth}
        \centering
        \includegraphics[width=\linewidth, height=\linewidth, keepaspectratio=false]{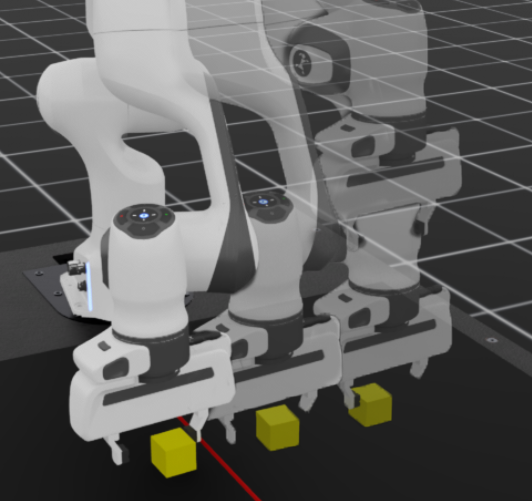}
        \vspace{-0.7cm}
        \par\medskip \small{(e) Push (Sim)}
    \end{minipage}
    \caption{Overview of evaluation tasks. The Pick and Place tasks (a,c) involve reaching the block (or mug), grasping it, lifting and placing it beyond a target line (or on the plate). The Push tasks (b,d) involve reaching the block and pushing it beyond a target line.}
    \label{fig:experiment_tasks}
\end{figure*}

\FloatBarrier %
\subsection{Learning Curves and Ablations}\label{app:learning-curves}

\begin{figure*}[!htbp]
    \centering
    \includegraphics[width=0.9\linewidth]{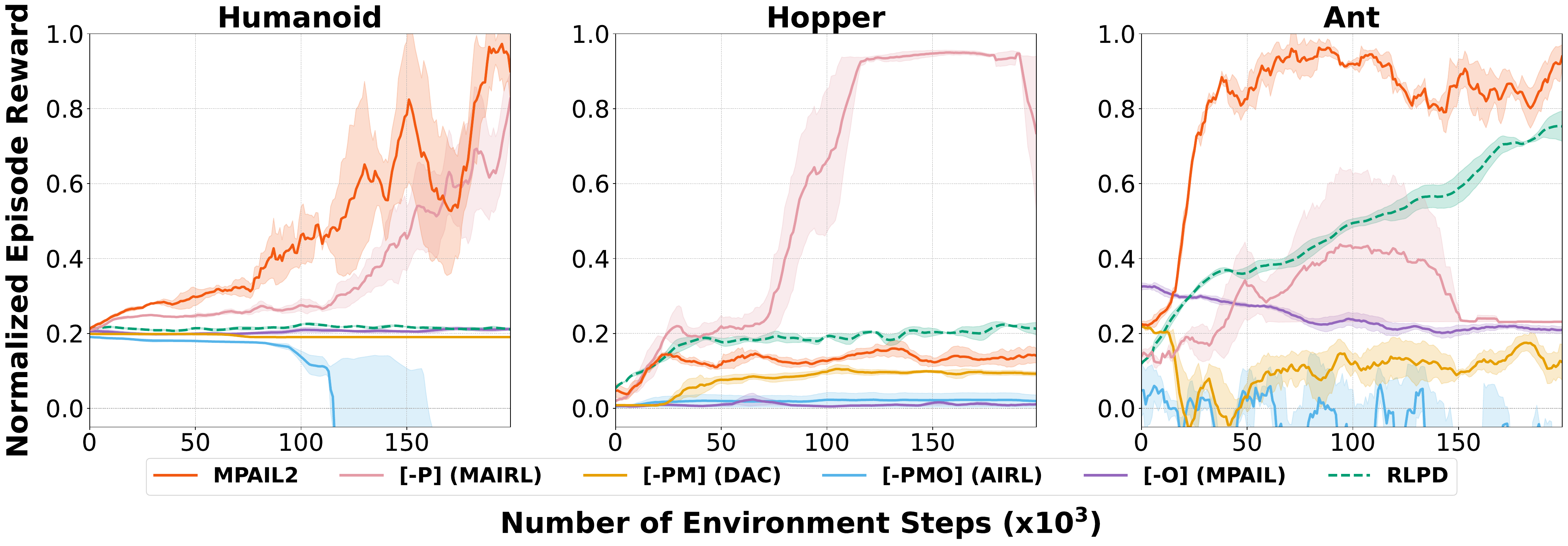}
    \caption{\textbf{Results in Gymnasium environments~\cite{towers_gymnasium_2025}}. All methods are provided 50 synthetic demonstration observations. RLPD is additionally provided dense reward and demonstration actions. Reward is normalized such that 1.0 reflects the average reward across the demonstrations.}
    \label{fig:gym-envs}
\end{figure*}

\begin{figure*}[!htbp]
    \centering
    \includegraphics[width=0.9\linewidth]{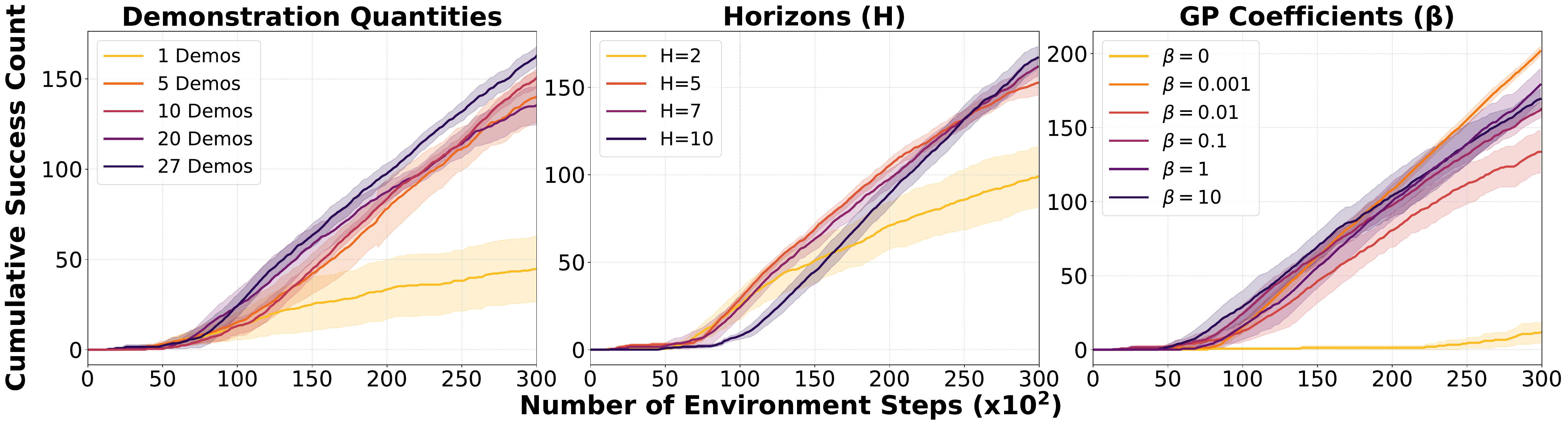}
    \caption{\textbf{\name ablations} over number of demonstrations (left), horizon lengths (middle), and gradient penalty coefficients (right) in Sim: Block Push.}
    \label{fig:abl-demo}
\end{figure*}

\begin{figure*}[!htbp]
    \centering
    \includegraphics[width=0.9\linewidth]{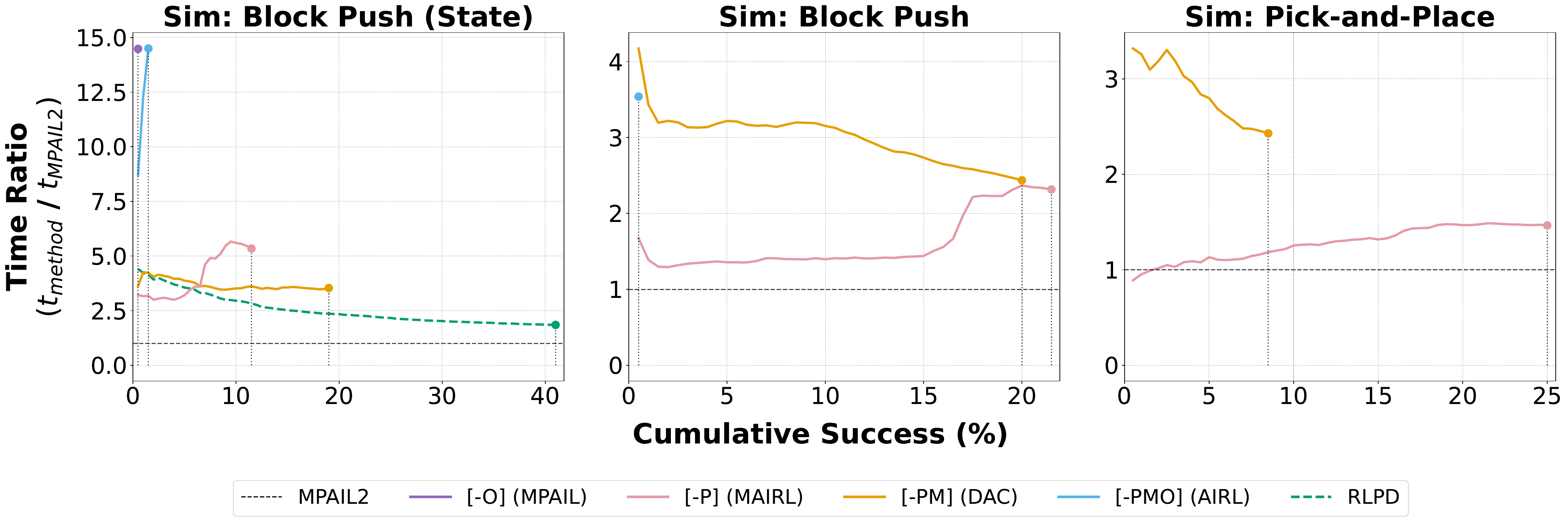}
    \caption{\textbf{Time efficiency of \name compared to other methods}. We plot $t_{\text{algo}}(c)/ t_{\text{MPAIL2}}(c)$ where $t(c)$ is the learning iteration at which the algorithm achieves a cumulative success percentage. With respect to a given baseline, the value reflects how many times faster \name achieves a given success percentage. An endpoint indicates where the corresponding baseline achieves its maximum success percentage.}
    \label{fig:time-efficiency}
\end{figure*}

\begin{figure*}[!htbp]
    \centering
    \includegraphics[width=0.9\linewidth]{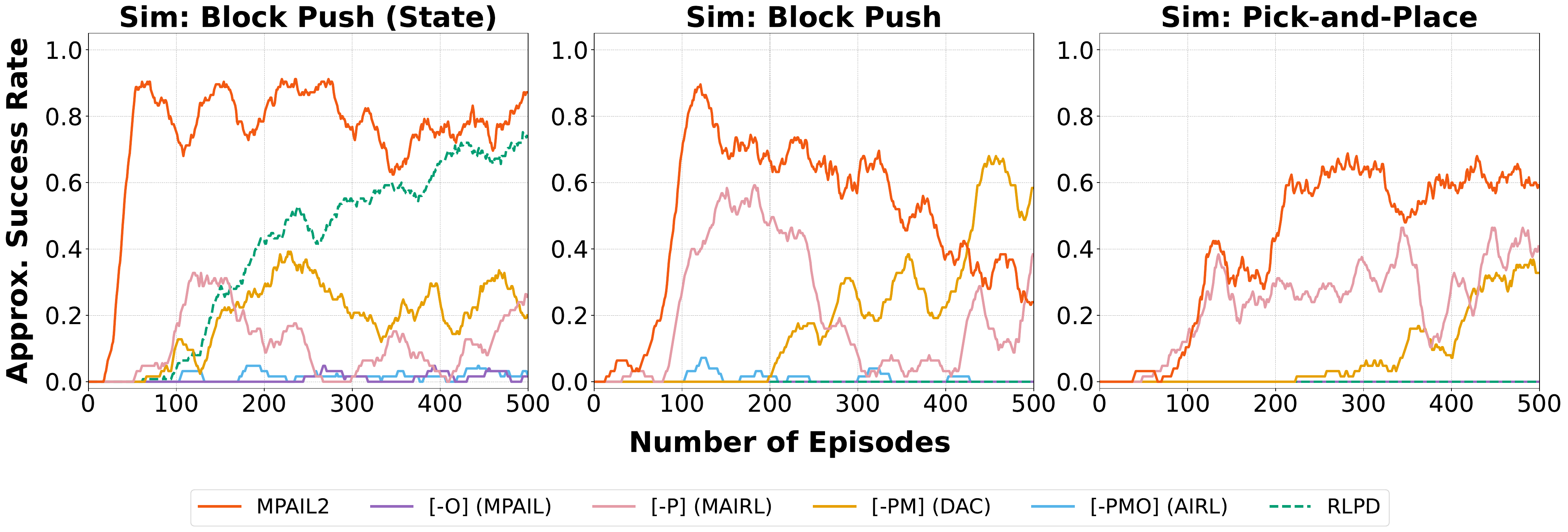}
    \caption{\textbf{Success rate stability}. We plot the average success rate of each method over a moving window of 25 episodes. A value of 1 indicates success on all episodes across 25 updates. This graph captures a method's stability across updates. In \textit{Sim: Block Push (State)} and \textit{Sim: Pick and Place}, \name is able to maintain a consistent success rate of roughly 0.8 and 0.6 respectively while the decreasing curve in \textit{Sim: Block Push} indicates instability as witnessed by the flattening of the cumulative success curve in Figure~\ref{fig:sim-results}. }
    \label{fig:success-rate}
\end{figure*}

\Cref{fig:gym-envs,fig:abl-demo,fig:time-efficiency,fig:success-rate} report additional learning curves in Gymnasium environments, ablations over key hyperparameters, sample efficiency relative to baselines, and per-episode success rate stability.

\FloatBarrier
\subsection{Analyses of: Predictability, Propagation, and Robustness}\label{app:sim-ablations}

These analyses probe the intuitions behind \name's design (\Cref{sec:intro}). Evaluations are performed on Sim: Block Push from images.

\begin{figure}[!htbp]
\centering
\includegraphics[width=\linewidth]{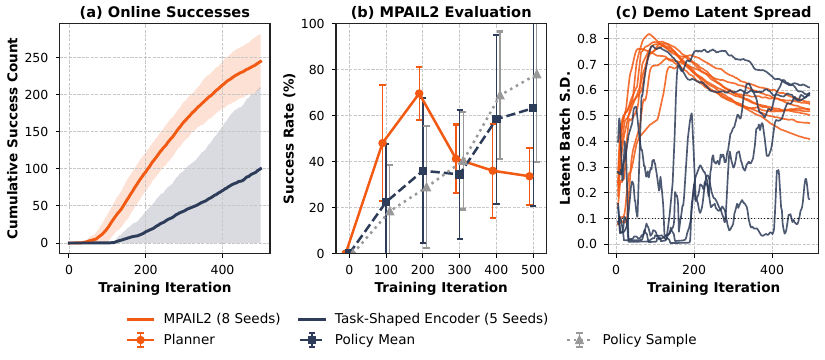}
\caption{\textbf{Simulated ablations}, Sim: Block Push (image). (a) Cumulative successes (mean $\pm$ std over seeds) of \name and of the variant whose encoder also receives the reward and value prediction loss gradients (i.e. TD-MPC2). (b) The same \name checkpoints executed with the planner, the policy's mean action, or a sampled policy action (mean $\pm$ std over 5 seeds; \Cref{tab:planner-onoff}). (c) Batch spread of the demonstration latents over training for every seed of both arms (10-iteration moving average); dotted: the collapse threshold 0.1.}
\label{fig:sim-ablations}
\end{figure}

\textbf{Task gradients into the encoder (predictability).} TD-MPC2~\cite{hansen_td-mpc2_2023} shapes its latent with reward and value losses in addition to latent prediction. We test this design under an inferred reward by also passing the gradients of the reward and value losses into \name's encoder, rescaled at every update to the size of the latent-prediction gradient so that neither dominates. The latent then tends to collapse: in 4 of 5 seeds the demonstration latents shrink to nearly a single point for part of training, which never happens to \name (\Cref{fig:sim-ablations}c). While collapsed, the reward can no longer tell demonstrations from the robot's own data and gives no learning signal. The 3 seeds that stay collapsed longest barely succeed, and on average cumulative successes fall by more than half, from 244 to 100 (\Cref{fig:sim-ablations}a). Keeping task gradients out of the latent, as \name does, avoids this failure. \label{app:tdmpc2-latent}

\textbf{Planner on vs.\ off (robustness).} We evaluate the same \name checkpoints with the planner and with the policy alone (\Cref{tab:planner-onoff}, \Cref{fig:sim-ablations}b). Early in training, executing the planner succeeds far more often than executing the policy, in every seed and by about 30 percentage points. Late in training the order flips, and the policy alone does better in 4 of 5 seeds. The policy that \name trains on its planner's data thus becomes competent on its own: the planner matters most early, while the policy still lags the reward, and the late reversal matches the decline of online success in \Cref{fig:sim-results}.

\begin{table}[!htbp]
\centering
\small
\begin{tabular}{lcccccc}
\toprule
Checkpoint (iteration) & 0 & 100 & 200 & 300 & 400 & 500 \\
\midrule
Planner & 0\,$\pm$\,0 & 48\,$\pm$\,25 & 70\,$\pm$\,11 & 41\,$\pm$\,15 & 36\,$\pm$\,20 & 34\,$\pm$\,12 \\
Policy mean, planner off & 0\,$\pm$\,0 & 22\,$\pm$\,25 & 36\,$\pm$\,32 & 34\,$\pm$\,28 & 58\,$\pm$\,37 & 63\,$\pm$\,43 \\
Policy sample, planner off & 0\,$\pm$\,0 & 18\,$\pm$\,20 & 29\,$\pm$\,26 & 40\,$\pm$\,21 & 69\,$\pm$\,28 & 78\,$\pm$\,38 \\
\bottomrule
\end{tabular}
\caption{\textbf{Planner on vs.\ off}, Sim: Block Push (image). Success rate (\%, mean $\pm$ std over 5 \name seeds, 50 episodes each) of the same checkpoints executed with the planner or with the policy alone.}
\label{tab:planner-onoff}
\end{table}

\begin{figure}[!htbp]
\centering
\includegraphics[width=\linewidth]{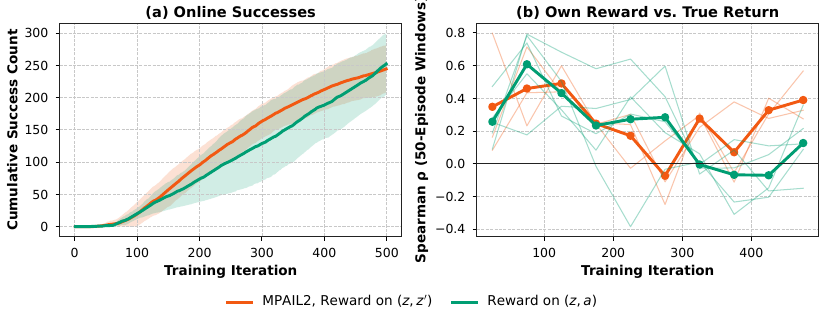}
\caption{\textbf{Reward on $(z, z')$ vs.\ $(z, a)$}, Sim: Block Push (image). (a) Cumulative successes (mean $\pm$ std over seeds). (b) Spearman correlation, within 50-iteration windows, between the run's own reward of each newly collected episode and that episode's true return (thin: seeds; thick: mean). The $(z, z')$ probe runs are the 3 \name seeds that logged it.}
\label{fig:coverage}
\end{figure}

\textbf{Reward on state and action.} Is there a difference in supervising outcomes rather than actions? We replace the reward's input, the observed transition $(z, z')$, with the state and commanded action $(z, a)$; for the demonstrations, which have no actions, the action is inferred from the end-effector's motion. Success is unchanged (\Cref{fig:coverage}a), but the two rewards differ in what they track. As each run collects episodes, we check whether its own reward ranks them in the order of their true return (\Cref{fig:coverage}b). Early in training both rewards do. Late in training only the observation reward still does; the state-action reward ranks the robot's episodes no better than chance (rank correlation 0.27 vs.\ 0.00). Once the policy acts like the demonstrations, its actions no longer separate good episodes from bad ones, while the observed outcome still does. On this task, that remaining signal does not translate into more successes.

\begin{figure}[!htbp]
\centering
\includegraphics[width=\linewidth]{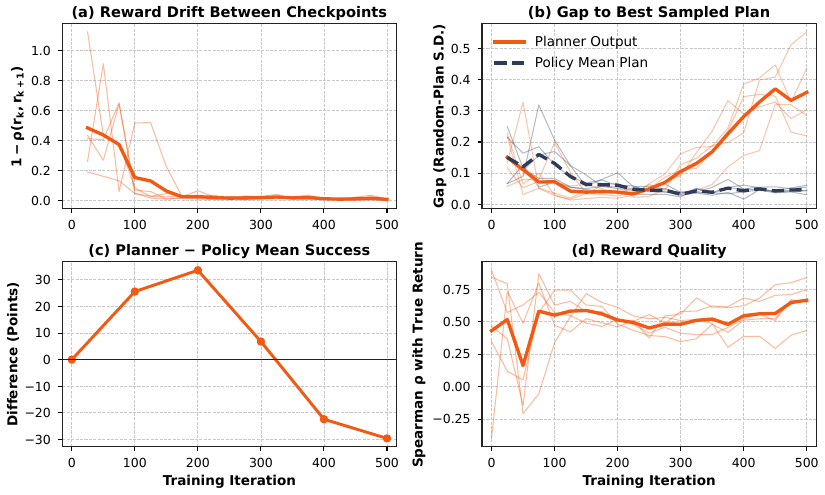}
\caption{\textbf{Reward drift and tracking}, Sim: Block Push (image), 5 \name seeds (thin) and mean (thick). (a) One minus the Spearman correlation between the rewards of checkpoints 25 iterations apart on a fixed probe set. (b) Score gap to the best sampled plan, under the planner's objective at each checkpoint, for the planner's output and the policy's mean plan, in units of the score s.d.\ of random plans. (c) Success of the planner minus success of the policy's mean action at the same checkpoints (\Cref{tab:planner-onoff}). (d) Spearman correlation between each checkpoint's summed reward and the true return over the 30 probe episodes.}
\label{fig:drift}
\end{figure}

\textbf{Reward drift and tracking (propagation and robustness).} To see how fast the reward changes, we score one fixed set of transitions with the reward of every checkpoint and measure how much its ranking of them changes between checkpoints. The reward changes quickly early in training and then settles (\Cref{fig:drift}a). To see how well each actor keeps up, we score the policy's plan and the planner's output with the planner's own objective at each checkpoint and compare both with the best plan the planner sampled (\Cref{fig:drift}b). While the reward moves, the policy falls behind the best plan, further the faster the reward moves, and the planner stays closer to it. Once the reward settles, the policy catches up, while the planner's averaged output drifts away from its own best sample. The two gaps cross shortly before the point where executing the planner stops beating executing the policy (\Cref{fig:drift}c). The reward's agreement with the true return does not drop late in training (\Cref{fig:drift}d), so the late reversal comes from the planner's search rather than from a worse reward. Plans are scored offline, without the planner's online warm start.

\FloatBarrier
\subsection{World Model Analysis}\label{app:world-model-figs}

These analyses probe latent predictability on real data (\Cref{sec:discussion}).

\begin{figure}[h]
\centering
\begin{minipage}[c]{0.56\linewidth}
\centering
\includegraphics[width=\linewidth]{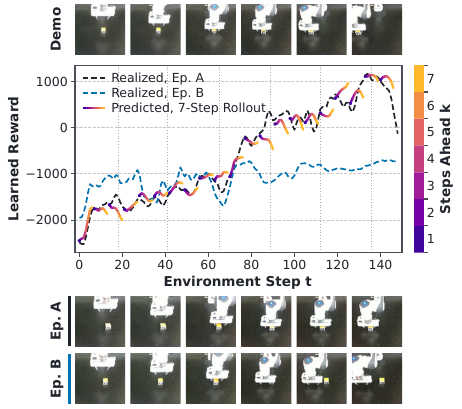}
\end{minipage}\hfill
\begin{minipage}[c]{0.41\linewidth}
\caption{\textbf{Held-out world-model check, real Block Push.} Ep.~A (episode 107) and Ep.~B (109) were collected after the checkpoint shown (iteration 100). \textit{Dashed}: reward $r(z_t,z_{t+1})$ of the encoded frames along each executed trajectory. \textit{Colored}: reward predicted by rolling the dynamics $k=1,\dots,7$ steps along the executed actions from every seventh state of Ep.~A. Ep.~A pushed the cube 22.8\,px toward the goal; in Ep.~B it never moved. Filmstrips: table camera at the dotted times; the demonstration (top) at the same fractions of its length.}
\label{fig:wm}
\end{minipage}
\end{figure}

\begin{figure}[!htbp]
\centering
\includegraphics[width=\linewidth]{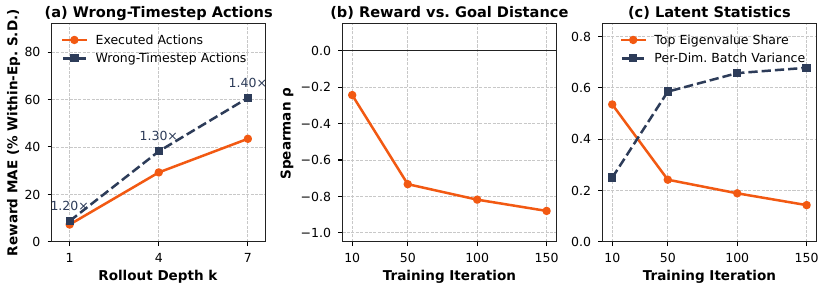}
\caption{\textbf{World-model diagnostics, first real Block Push run} (the run of \Cref{fig:wm}). (a) Reward-prediction error through $k$-step latent rollouts (99 held-out episodes, iteration 100), along the executed actions and along actions from a different timestep of the same episode; labels give the ratio. Only $k=1$ approximates a causal test (see text). (b) Spearman correlation between episode-mean reward and final cube-to-goal distance on 49 episodes later than every checkpoint shown. (c) Latent statistics on the same episodes as (b).}
\label{fig:wm-diag}
\end{figure}

\begin{figure}[!htbp]
\centering
\includegraphics[width=\linewidth]{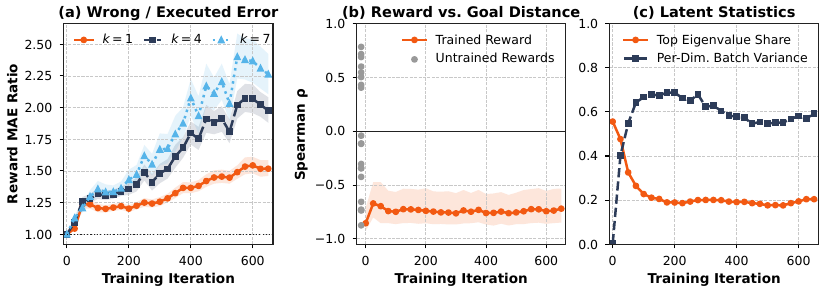}
\caption{\textbf{The same diagnostics at every 25th checkpoint of a second real Block Push run}, on its last 49 episodes (held out for checkpoints up to 650). (a) Ratio of $k$-step reward-prediction error with wrong-timestep actions to that with executed actions (1.0 = the model ignores the action; untrained models give 0.98--1.01). (b) Spearman correlation of episode-mean reward with final cube-to-goal distance; gray dots are 20 untrained reward heads. (c) Latent statistics.}
\label{fig:wm-diag-run2}
\end{figure}

\begin{figure}[!htbp]
\centering
\includegraphics[width=\linewidth]{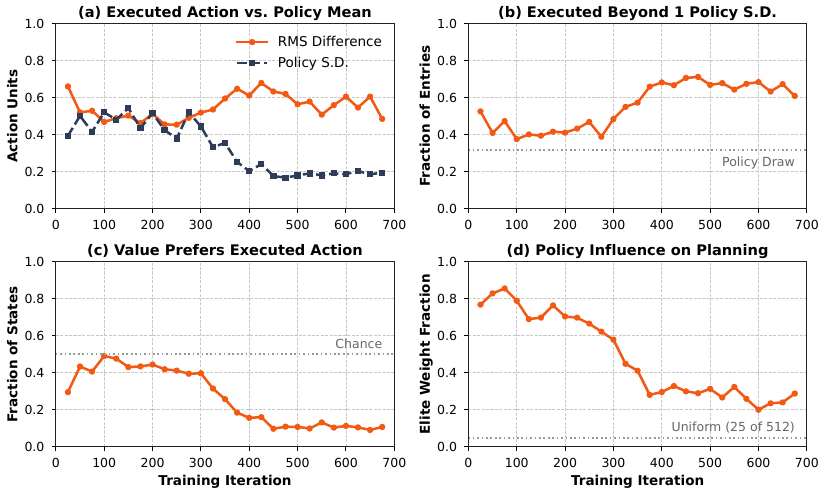}
\caption{\textbf{Planner vs.\ policy on the second real Block Push run.} At each checkpoint, the executed planner action is compared with the policy's mean action at the same state. (a) RMS difference in action units, with the policy's standard deviation for scale. (b) Fraction of (step, dimension) entries where the executed action lies more than one policy standard deviation from the policy mean (dotted: 0.317, the fraction for a draw from the policy). (c) Fraction of states where the 5-head mean value prefers the executed action to the policy mean (dotted: chance). (d) Policy influence (\Cref{fig:pol-influence}): the share of the final MPPI iteration's elite weight on the 25 policy-seeded samples, from re-planning offline on the stored observations (dotted: 25/512, the share if they were no better than the other samples).}
\label{fig:planner-vs-policy}
\end{figure}

\textbf{Held-out diagnostics on real Block Push.} We run three diagnostics offline on the replay buffer of the run in \Cref{fig:wm} (\Cref{fig:wm-diag}) and repeat them at every 25th checkpoint of a second, longer run on the same task that is not one of the paper's seeds (\Cref{fig:wm-diag-run2,fig:planner-vs-policy}). Each checkpoint is evaluated only on episodes collected after it.

\textit{Action dependence.} Rolled out along the executed actions, the dynamics predicts the learned reward closely one step ahead, and the error grows from 7\% to 43\% of the reward's spread within an episode between one and seven steps ahead, the planning horizon. Early in training, simply repeating the last reward predicts better beyond one step; the model overtakes this baseline as training proceeds. To test whether the prediction uses the action, we replace the executed action with an independent planner sample from the same state: one-step error rises by about 9\%, so the dynamics uses the action's effect, but modestly. Swapping in actions from other timesteps raises the error more at longer horizons (\Cref{fig:wm-diag-run2}a), but on closed-loop data those actions also carry information about future states, so only the one-step test is close to causal.

\textit{Reward vs.\ task progress.} We correlate each held-out episode's mean reward with how far the cube ended from the goal. Training makes this correlation strongly negative, as it should be: the closer the cube ends to the goal, the higher the reward (\Cref{fig:wm-diag}b). Untrained reward networks, however, reach correlations of similar size with random sign, since almost any function of the table image tracks where the cube ended up (\Cref{fig:wm-diag-run2}b). Training fixes the sign, which is correct at every trained checkpoint; the size of the correlation alone is not evidence that the reward tracks task progress.

\textit{Collapse.} The latent does not collapse on real data: the share of its variance along a single direction falls from about half to under a fifth as training proceeds (\Cref{fig:wm-diag}c). The stop-gradient target, multi-step prediction, low encoder learning rate, and output LayerNorm keep it away from the constant solution that a zero prediction loss would allow~\cite{chen_exploring_2021, tang_understanding_2023}, though it occupies only about a dozen effective dimensions of 512.

\textit{Planner vs.\ policy.} On the second run, the planner's executed action stays away from the policy's mean action, even after the policy narrows late in training (\Cref{fig:planner-vs-policy}a,b). Early on, the planner usually executes a plan seeded by the policy; later it mostly finds better plans of its own (\Cref{fig:planner-vs-policy}d). Late in training the value prefers the policy's mean action in most states (\Cref{fig:planner-vs-policy}c): the planner optimizes the model-based return over its horizon, while the policy maximizes a target that is mostly the value's bootstrap (\Cref{eq:lam-return}). The elite weights are effectively one-hot at these return scales, so the executed action is the best sampled plan.

\begin{figure}[t]
    \centering
    \includegraphics[width=.7\linewidth]{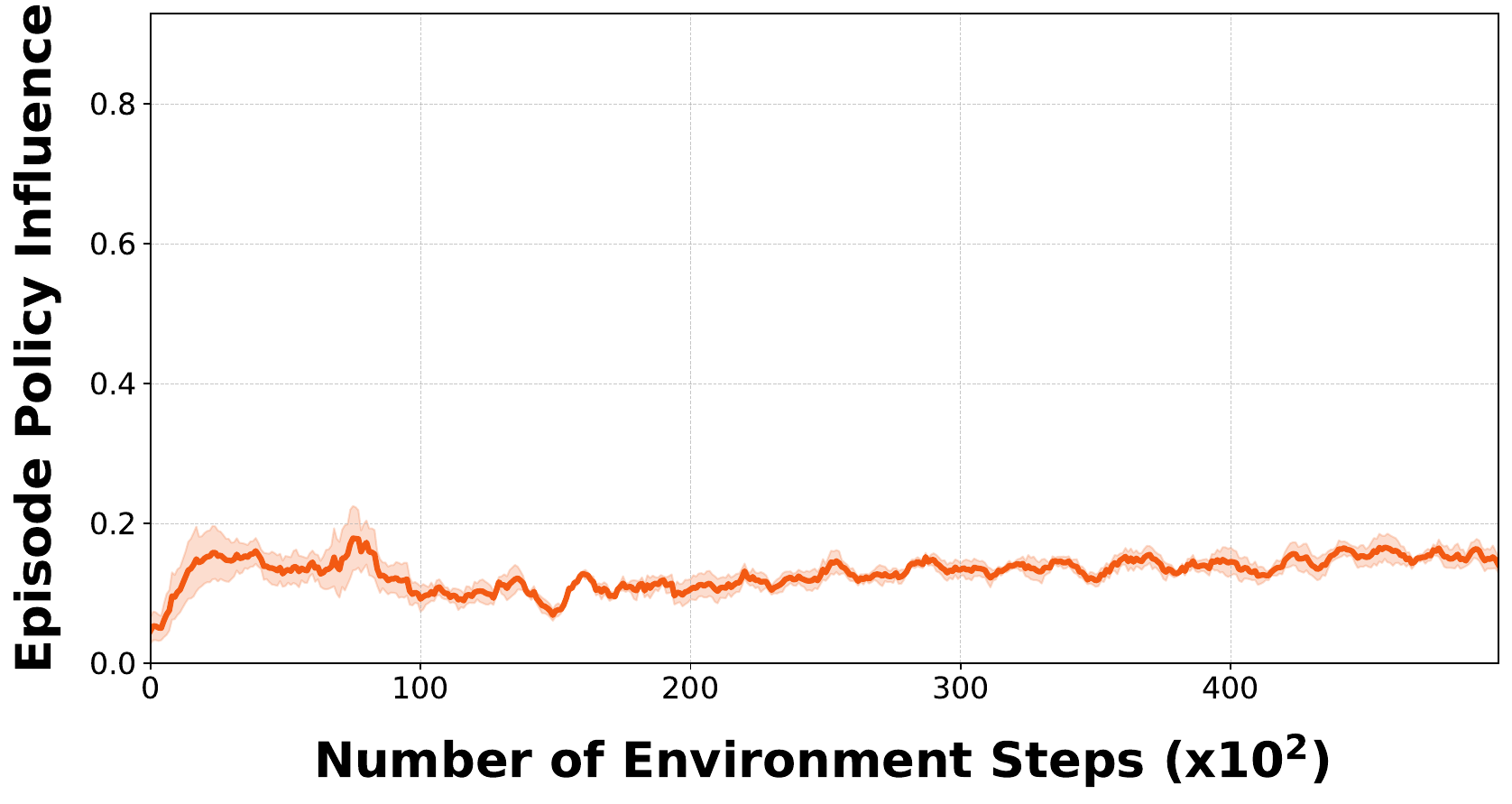}
    \caption{\textbf{Policy influence on planning.} Influence is computed as the fraction of total MPPI score attributable to policy-seeded rollouts among elite samples at the final planning iteration (see~\Cref{alg:mppi-procedure}), averaged over all timesteps in an episode.}
    \label{fig:pol-influence}
\end{figure}

\Cref{fig:pol-influence} quantifies how much the explicit policy contributes to MPPI planning scores over training, revealing that the learned policy contributes relatively little weight to the final optimized online plan.

\begin{figure}[t]
    \centering
    \includegraphics[width=0.72\linewidth]{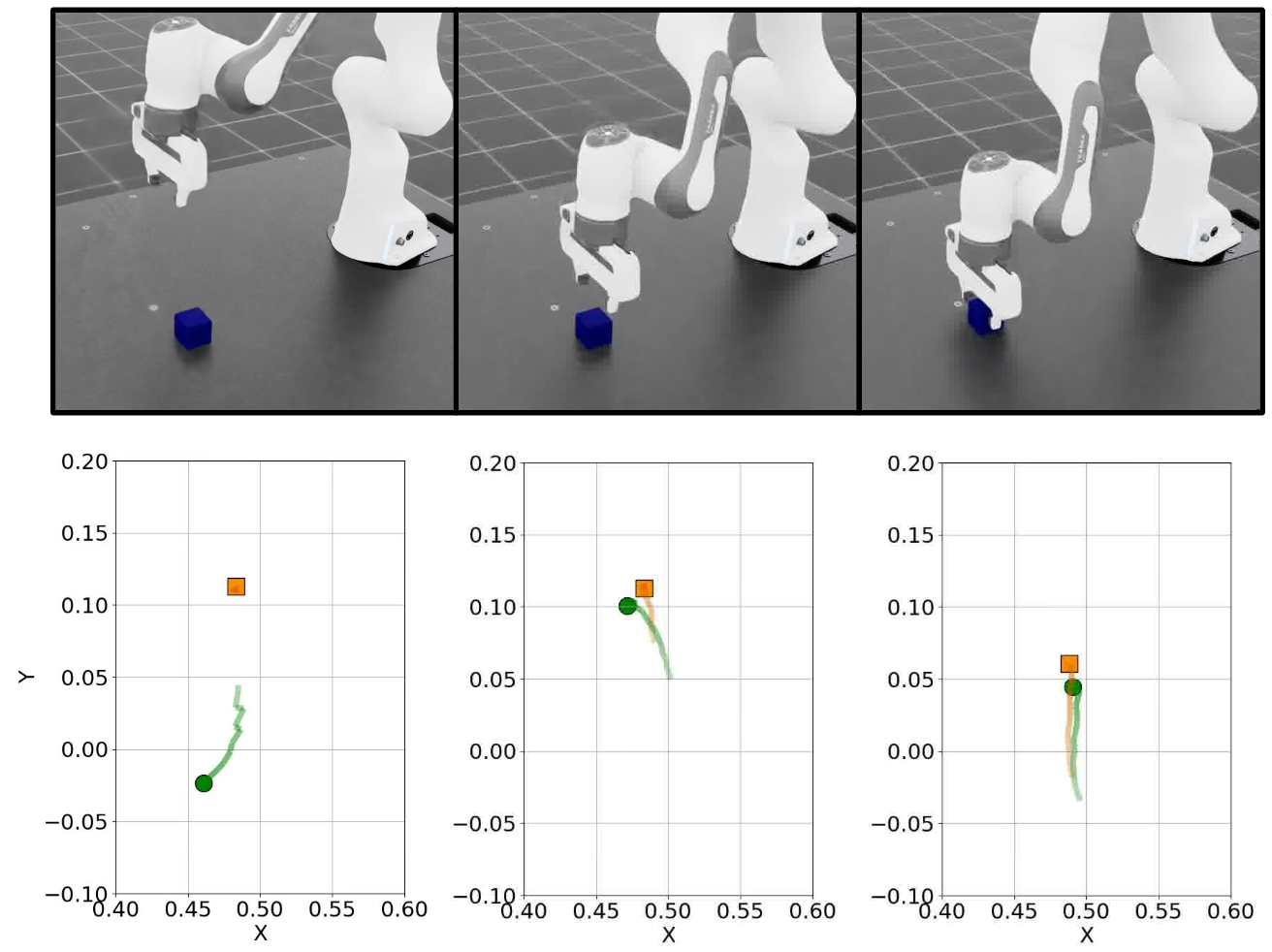}
    \caption{\textit{Top.} \name on Sim: Block Push (state) at three timesteps. \textit{Bottom.} Predicted plans in the XY plane for the next one second. End-effector plans in \textcolor{OliveGreen}{green}; block plans in \textcolor{YellowOrange}{orange}. Block trajectories are static until contact, illustrating causal world modeling. \textit{Planning is in latent space; a separately trained decoder is used for visualization.}}
    \label{fig:small-vis}
\end{figure}

\Cref{fig:small-vis} visualizes the planned rollouts in the Sim: Block Push (state) task where the learner's prediction of the block trajectory can be displayed. The block's static trajectory becoming more dynamic as the robot approaches provides evidence towards \name's implicitly causal structure.

\begin{figure}
    \centering
    \includegraphics[width=\linewidth]{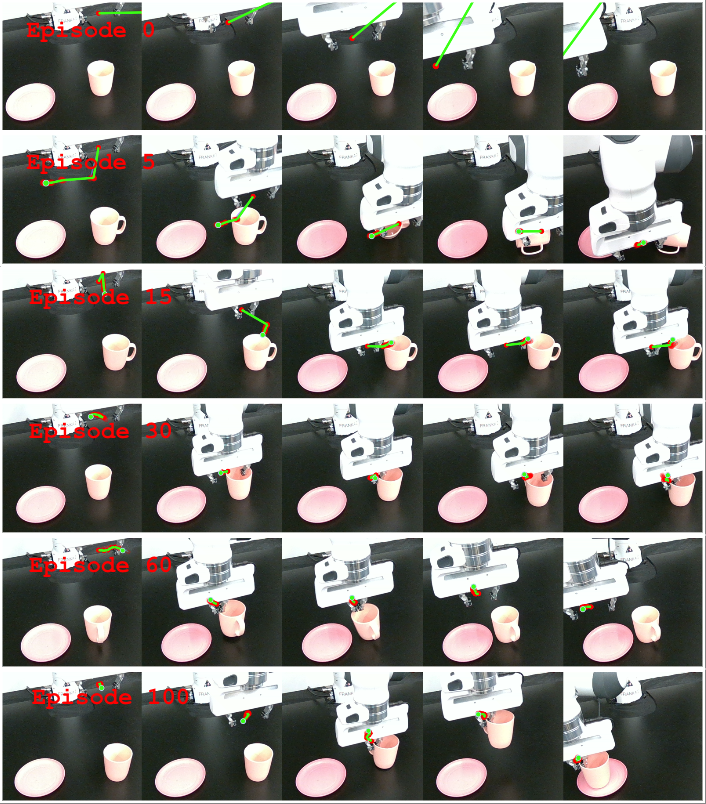}
    \caption{Real-world planning visualization on the Mug-on-Plate task. Each row shows five frames at training iterations 0, 5, 15, 30, 60, and 100. \textcolor{red}{Red} lines show the top-$k$ end-effector plans; the \textcolor{green}{green} line is the optimal plan selected by the planner. Early episodes show unstructured, unmodeled, and exploratory behavior; by episode~100 the robot reaches, grasps, and places the mug onto the plate.}
    \label{fig:mop-vis}
\end{figure}

\textbf{\name's latent plans can be reconstructed for inspection and verification.} We find that the model correctly learns over time that block motion can only be influenced through contact (\Cref{fig:small-vis}). When performed online, these visualizations can help practitioners identify where (dynamics, reward, value) the agent may benefit from more supervision or training. These visualizations provide clear qualitative evidence that the robot's world model is improving over time, as seen in \Cref{fig:mop-vis}.

\section{Additional Discussion}\label{app:discussion}

The following discussions provide additional evidence, details, observations, and summaries beyond the main body.

\textit{RLPD metacommentary:} We emphasize that RLPD is competently implemented; given state observations, dense reward, and action-labeled demonstrations, it is the second strongest method on Sim: Block Push (State). Its difficulty is concentrated in the visual, real-world setting, where -- within the small interaction budgets considered here, and despite the additional reward and action supervision it requires -- it does not yet exhibit successes. This pattern is consistent with the well-known sample complexity of visual RL from scratch rather than an intrinsic failure of the method.
We also note that prior work reports only wall-clock time and does not include the number of interactions collected~\cite{luo_serl_2025}. While wall-clock times for pick-and-place appear similar in magnitude (hours) to prior work, our setups are not engineered to optimize real-world training time (relevant future work); we therefore expect that the real-world RLPD results in~\cite{luo_serl_2025} collect more interactions over the same period, and that RLPD might improve given more interaction. Accordingly, our comparison should be read as a statement about sample efficiency under matched, low interaction budgets and minimal supervision, rather than about the asymptotic capability of RLPD. However, the IRL baselines' higher efficiency under strictly less supervision challenges whether asymptotic performance is practical in the context of real-world learning.

\subsection{Why Is \name\ Significantly More Sample Efficient? Building up from baselines.}\label{app:disc-efficiency}

This discussion proceeds by incremental analyses from RLPD$\rightarrow$DAC$\rightarrow$MAIRL$\rightarrow$MPAIL2. BC and MPAIL(1) comparison insights are discussed following.

\mypara{RLPD$\rightarrow$DAC.}
Unlike RLPD, DAC demonstrates online improvement, albeit much slower than \name, by learning to reach and sometimes interact with the object during training.
The primary architectural advancement of DAC over RLPD is the online reward model, instead of the hand-designed reward. In the event of using a classifier-based reward for RLPD, then the difference somewhat reduces to reward regularization (i.e. DAC utilizes the Gradient Penalty~\cite{gulrajani_improved_2017} with the exception that classification in RLPD remains actively supervised online by human labeling. 

It remains possible that real-world RLPD depends heavily on pre-training of the reward and encoder~\cite{luo_serl_2025}, neither of which was included in our experiments. Such components rely on additional prior information and supervision (curated failure examples for the reward classifier and large-scale encoder pre-training) that, for a fair comparison, were likewise withheld from the other methods. We additionally ran RLPD out to 300 episodes on Block Push ($2\times$ the standard budget) and observed no improvement in behavior. Despite RLPD's nominal ability to train from scratch, its real-world demonstrations have been evaluated almost exclusively with pre-trained visual encoders~\cite{luo_serl_2025, luo_precise_2025}. We regard the absence of a pre-trained, label-supervised encoder less as a constraint than as consistent with a long-standing position that visual representations for an embodied agent should emerge from real-world interaction with natural input rather than from curated external supervision~\cite{olshausen_emergence_1996, bajcsy_revisiting_2018, lecun_path_nodate}.

\mypara{DAC$\rightarrow$MAIRL.}
The addition of the encoder and dynamics model significantly improves learning efficiency to the extent where numerous successes are exhibited. Until this advancement, under RLPD and DAC, the robot spends most training episodes ``over-reaching'' by moving to the edges of the workspace repeatedly. With MAIRL, this behavior ends far earlier in training, reflecting much more efficient learning. As noted in~\Cref{app:baselines}, an encoder was also provided to DAC, trained via gradients from the value target loss. Thus, the improvements in this step lie in the supervision of the latent space (i.e. dynamics) and the policy optimization objective.

\textbf{MAIRL$\rightarrow$MPAIL2.}
Finally, the addition of the planner reduces training times such that real-world training becomes practical and performance consistently improves upon BC, despite BC's privileged supervision. More surprising is the planner's ability to enable transfer learning where MAIRL and BC exhibit negative transfer. In this advancement, recovery behavior is more robust and appears far earlier than MAIRL.

\textbf{BC (Diffusion).}
We observed that BC often exhibited behavior imitating exactly the demonstration data. This appears to work in its favor for Real: Block Push. While the task is dynamically sensitive, block destabilizations often remain in-distribution to the demonstration data. Thus, the BC policy is still able to draw upon demonstration coverage for task success. We observe that BC is reliable in resets that are ``in the convex hull'' of the demonstrations. Placing block, mug, or plate outside this ``hull'' region often results in failure for BC.

This rote memorization and interpolation works against the BC policy for Pick-and-Place as it results in a success rate of only $12\%$ compared to \name's $82\%$. In many failure cases, the BC policy does not grasp the cube but still continues to execute the remainder of the demonstration trajectory without the block grasped. In contrast, \name occasionally misses the grasp but often returns to the block to reattempt a grasp. More generally, \name often appears to exhibit recovery behaviors much earlier in training than other online baselines. While Diffusion Policy has been previously shown to exhibit recovery behaviors, it does not appear to do so in our experiments, likely due to the limited data regime.

\textbf{MPAIL.}
The poor performance of \nameo{} should not be interpreted as evidence that planning or latent dynamics are insufficient. Rather, \nameo{} preserves both components and
differs primarily in the use of on-policy reward, value, and policy updates. Under the matched low-interaction budgets considered here, the on-policy rollout buffer provides
substantially narrower state-transition coverage than the accumulated replay buffer used by \name{}. This makes the adversarial reward more prone to rapidly fitting the current
learner distribution and gives the value/policy updates fewer diverse transitions from which to learn recovery behavior. The result is that, despite retaining the planner and world model, \nameo{} does not match the sample efficiency or stability of \name{}.

\subsection{Discussion by Task}\label{app:disc-task}

\textit{Block Push:} In simulation, Block Push presents as a more dynamically unstable task. This characterization might explain BC's results in~\Cref{tab:results}, generally performing worse than \namep, \namepm, and \name. Often, BC appears to stay near expert trajectories when the block destabilizes completely beyond the gripper. In the real-world, these dynamics are less unstable, likely due to higher friction and slower actuation, resulting in higher real-world BC performance.

Regarding higher training instability observed in Block Push for \name and \namep, a possible reason is that the expert never exhibits recovery from destabilization. However, as learners are always taking sampled, potentially suboptimal actions while learning, the learner almost always demonstrates destabilization and recovery. As a result, the inferred reward may become highly precise over training, making it increasingly difficult for the learner to meet the demands of the precise reward. The planner in \name likely helps mitigate losing track of rewards entirely through long-horizon planning, resulting in significantly more stable learning than \namep.
Nevertheless, both simulated and real-world results empirically support pursuing future work which might help stabilize adversarial rewards or establish practical stopping-criteria.

\textit{Pick-and-Place:} Compared to Block Push, Pick-and-Place is less dynamically sensitive. While it can be catastrophically unstable, we find that performance on Sim: Pick-and-Place can be achieved by depending on the demonstrations. BC's high relative success suggests that the task does not require significant generalization beyond the demonstrations. However, Real: Pick-and-Place appears to require greater generalizability with policy methods (\namep and BC) both dropping to below $16\%$ in real-world from $40\%$ in simulation. Observation noise, delays, and other real-world challenges are believed to contribute to the drop in performance. Planning appears to be robust against these challenges.

\textit{Mug-on-Plate:} To our knowledge, mug-on-plate diverges from current real-world RL-from-scratch evaluations in that there exists randomization in the object's target location, i.e. the plate. In other works like~\cite{sun_adversarial_2021, luo_serl_2025, luo_precise_2025}, the target object (e.g. USB socket, cam fittings, PCB mounting points) is usually locked down, allowing the robot to rely on the proprioception as the task objective. In more unstructured settings, locking down the target object may not be possible and hence the MoP task aims to challenge methods in this way. \name outperforms BC on average. We remark that \name also tends to precisely place the mug in the center of the plate, while BC tends to miss the precise center though we still mark these performances as successes.

\textit{Gymnasium Environments:} We evaluate \name and all the baselines on 3 environments from Gymnasium : \textit{Ant-v5, Hopper-v5, Humanoid-v5} (\Cref{fig:gym-envs}). We run each method for 200 training iterations with only 1 environment to mimic real-world training settings where parallelization is not possible. Due to the high-action spaces of Humanoid (17) and Ant (8), these environments validate the role of the policy in seeding online plans. Prior work in MPAIL demonstrated poor performance on Ant without policy seeding.

We suspect the suboptimal performance of \name in Hopper is primarily due to the environment's highly restrictive terminations adversely affecting planning. With terminations enabled, the learner is unable to collect experiences that would otherwise present counterfactual experience critical to planning. Future work assuming termination knowledge may benefit from providing this information to the learned dynamics model.

\subsection{Problem Learning Intuition}\label{app:problem-learning}

As posited by prior work, MPAIL can be viewed as requiring the agent to learn a problem in addition to its solution, where the problem is given by the components: dynamics, reward, and value; and the problem's solution is given by the policy. This perspective provides intuition about why explicit policies can perform worse than planners (or iterative methods, in general~\cite{pan_much_2025}) despite optimizing equivalent objectives over similar data.

This intuition appears consistent in our visual manipulation experiments. \name achieves high task performance with a small policy fraction (M/N = 0.05), and throughout Sim: Block Push the policy's contribution to the elite planning score remains below $20\%$~(\Cref{fig:pol-influence}), increasing slowly relative to task progress. The causal role of planning is further corroborated by the ablation: removing planning entirely ({[}$-$P{]}, MAIRL) substantially degrades performance across all tasks~(\Cref{tab:results}). Together these support the rationale from~\Cref{sec:method-policy} that the explicit policy primarily seeds online optimization and supports off-policy value estimation, rather than serving as the primary execution mechanism.

Our results corroborate general efforts in iterative (implicit policy) methods as a generalizable alternative to explicit policies~\cite{jain_smooth_2025, chi_diffusion_2025, pan_much_2025}. Consequently, these implicit models obey an iteration objective. In Diffusion, in-distribution (energy-based) score is the iteration objective. In the case of \name, the model-based return is the iteration objective. We refer readers to~\cite{li_unifying_2025} for an in-depth treatment of these equivalences.

\subsection{Why Does \name\ Exhibit Transfer?}\label{app:disc-transfer}

A long-standing challenge of continual reinforcement learning is the counterintuitive decrease in performance when policies, pre-trained on prior tasks, are fine-tuned or continually adapted for new tasks or environments~\cite{rudin_parkour_2025}. Model-based representations have been believed to hold the key to unlocking continual improvement by rooting the learner's progression in the underlying dynamics of the world~\cite{sutton_reinforcement_2018}.
We believe the transfer experiment results in \Cref{fig:transfer-results} point to promising signs of these hypotheses operating in the real-world.

We reiterate that this work is the first to demonstrate online transfer learning between tasks (demonstrations) entirely in the real-world from scratch. Thus, prior work is relevant insofar as they encourage discussion as they differ in pre-training, tasks, and continual learning constraints.
Prior methods have approached transfer learning through methods in data buffers, architectures, residual models, meta-learning, and more~\cite{julian_never_2021, nagabandi_deep_2019, nagabandi_learning_2019}. By contrast, \name demonstrates transfer strictly through model-based representations and planning. Because \name must still be subject to effects of plasticity loss, we hypothesize that transfer is enabled by the ability to \textit{resolve new policies online}. As discussed in~\Cref{app:problem-learning}, this capability of planning allows the learner to ignore its prior ``solution'' (policy) if it does not adhere to the ``problem'' (dynamics, reward, value) of the new task.

We conduct an additional transfer experiment where only the encoder and dynamics models are transferred to validate the transferability of the latent representation.
Specifically, the weights of only $\enc$ and $\dyn$ are initialized using the pre-trained weights. The remaining components (reward $\rew$, value $\val$, policy $\pol$) are randomly initialized (trained from scratch). The results of this experiment can be found in \Cref{fig:transfer-results}. The results of \name indicate that the dynamics-only transfer capabilities remain about the same as the full transfer.

\name deserves more attention to its transfer capabilities than can be provided in just this work's evaluations. However, by taking significant steps towards real-world transferable IRLfO, we believe that this work forms a strong foundation for further, more complex, experiments.

\subsection{Areas of Improvement}\label{app:disc-improvement}

The visual manipulation experiments chosen in this work reflect a decision to trade off experiment complexity for statistical significance given the sparse adjacent work. Whether \name extends to longer-horizon, multi-object tasks, higher-resolution observations, or large datasets is not evaluated in this work; its transfer results, off-policy design~\cite{hansen_td-mpc2_2023}, and IRL's demonstrated scalability elsewhere~\cite{barnes_massively_2023, zakka_xirl_2022} make these natural next tests rather than established capabilities.

\mypara{Stabilization methods.} The adversarial manner in which the reward is trained can present as learning instability or inconsistency. In applications, it may be challenging or impossible to depend upon laboratory-setting metrics like reward or success rate which make early-stopping and deployment possible. The ability to continually train and improve monotonically with experience is necessary for real-world performance. 

\mypara{Uncertainty, exploration, and safety.} World models, as used in \name, help compartmentalize sources of uncertainty. In the case of \name, the encoder, dynamics, value, reward, policy, and planner are each responsible for different aspects of reasoning and acting. In this framework, it is more straightforward to rationalize what a learner might improve when the task is not the only objective.

\textbf{Action and reward assumptions.} While this work investigates more restrictive settings without access to demonstration action or hand-designed rewards, \name can be extended to leverage additional prior knowledge as in~\cite{peng_amp_2021} or in~\cite{ball_efficient_2023}.

High variability in first-success times in both simulation and real-world experiments indicates that supervision of the latent space deserves further attention. Pre-trained visual encoders, initialization methods, or auxiliary losses may help in reducing variability.

Planning visualizations during development (as in~\Cref{fig:small-vis}) provide insight into reasons for performance decreases or failures when they occur. As we have only visualized Sim: Block Push (state), the following conclusions may not be generally applicable. Nevertheless, these visualizations indicate that failures of \name are often caused by collapsed plans at the beginning of the episode. Rarely does \name fail while in the middle of the task. Rather, when they occur, failures often find the agent moving cyclically or back-and-forth.

\end{document}